\documentclass[10pt,twocolumn,letterpaper]{article}

\usepackage[pagenumbers]{cvpr} 

\usepackage[table,dvipsnames]{xcolor}
\usepackage{graphicx}
\usepackage{amsmath}
\usepackage{amssymb}
\usepackage{booktabs}
\usepackage[table]{xcolor}
\usepackage{pifont}
\usepackage{bm}
\usepackage{multirow}
\usepackage{rotating}
\usepackage[accsupp]{axessibility}
\definecolor{darkgreen}{RGB}{0, 150, 0}
\definecolor{darkred}{RGB}{200, 0, 0}
\definecolor{darkblue}{RGB}{0, 0, 200}
\newcommand{\ch}{{\color{darkgreen} \ding{51}}}
\newcommand{\xm}{{\color{darkred} \ding{55}}}

\usepackage[pagebackref,breaklinks,colorlinks]{hyperref}

\newif\ifdraft\drafttrue

\ifdraft

\else

\fi

\definecolor{gray9}{gray}{.9}
\definecolor{gray95}{gray}{.95}
\definecolor{gray8}{gray}{.8}
\definecolor{gray85}{gray}{.85}

\usepackage[capitalize]{cleveref}
\crefname{section}{Sec.}{Secs.}
\Crefname{section}{Section}{Sections}
\Crefname{table}{Table}{Tables}
\crefname{table}{Tab.}{Tabs.}

\def\cvprPaperID{4937} 
\def\confName{CVPR}
\def\confYear{2023}

\begin{document}


\title{DejaVu: Conditional Regenerative Learning to Enhance Dense Prediction}

\author{
Shubhankar Borse 
\and
Debasmit Das \thanks{These authors contributed equally to this work.}
\and
Hyojin Park \footnotemark[1]
\and
Hong Cai
\and
Risheek Garrepalli
\and
Fatih Porikli\\
{Qualcomm AI Research \thanks{Qualcomm AI Research, an initiative of Qualcomm Technologies, Inc.} }\\
{\tt\small \{sborse, debadas, hyojinp, hongcai, rgarrepa, fporikli\}@qti.qualcomm.com}\\
}

\maketitle

\begin{abstract}

We present DejaVu, a novel framework which leverages conditional image regeneration as additional supervision during training to improve deep networks for dense prediction tasks such as segmentation, depth estimation, and surface normal prediction. First, we apply redaction to the input image, which removes certain structural information by sparse sampling or selective frequency removal. Next, we use a conditional regenerator, which takes the redacted image and the dense predictions as inputs, and reconstructs the original image by filling in the missing structural information. In the redacted image, structural attributes like boundaries are broken while semantic context is largely preserved. In order to make the regeneration feasible, the conditional generator will then require the structure information from the other input source, i.e., the dense predictions. As such, by including this conditional regeneration objective during training, DejaVu encourages the base network to learn to embed accurate scene structure in its dense prediction. This leads to more accurate predictions with clearer boundaries and better spatial consistency. When it is feasible to leverage additional computation, DejaVu can be extended to incorporate an attention-based regeneration module within the dense prediction network, which further improves accuracy. Through extensive experiments on multiple dense prediction benchmarks such as Cityscapes, COCO, ADE20K, NYUD-v2, and KITTI, we demonstrate the efficacy of employing DejaVu during training, as it outperforms SOTA methods at no added computation cost.

\end{abstract}

\vspace{-10pt}
\section{Introduction}
\label{sec:intro}\vspace{-3pt}

\begin{figure}[t]
\begin{center}
\includegraphics[width=0.93\linewidth]{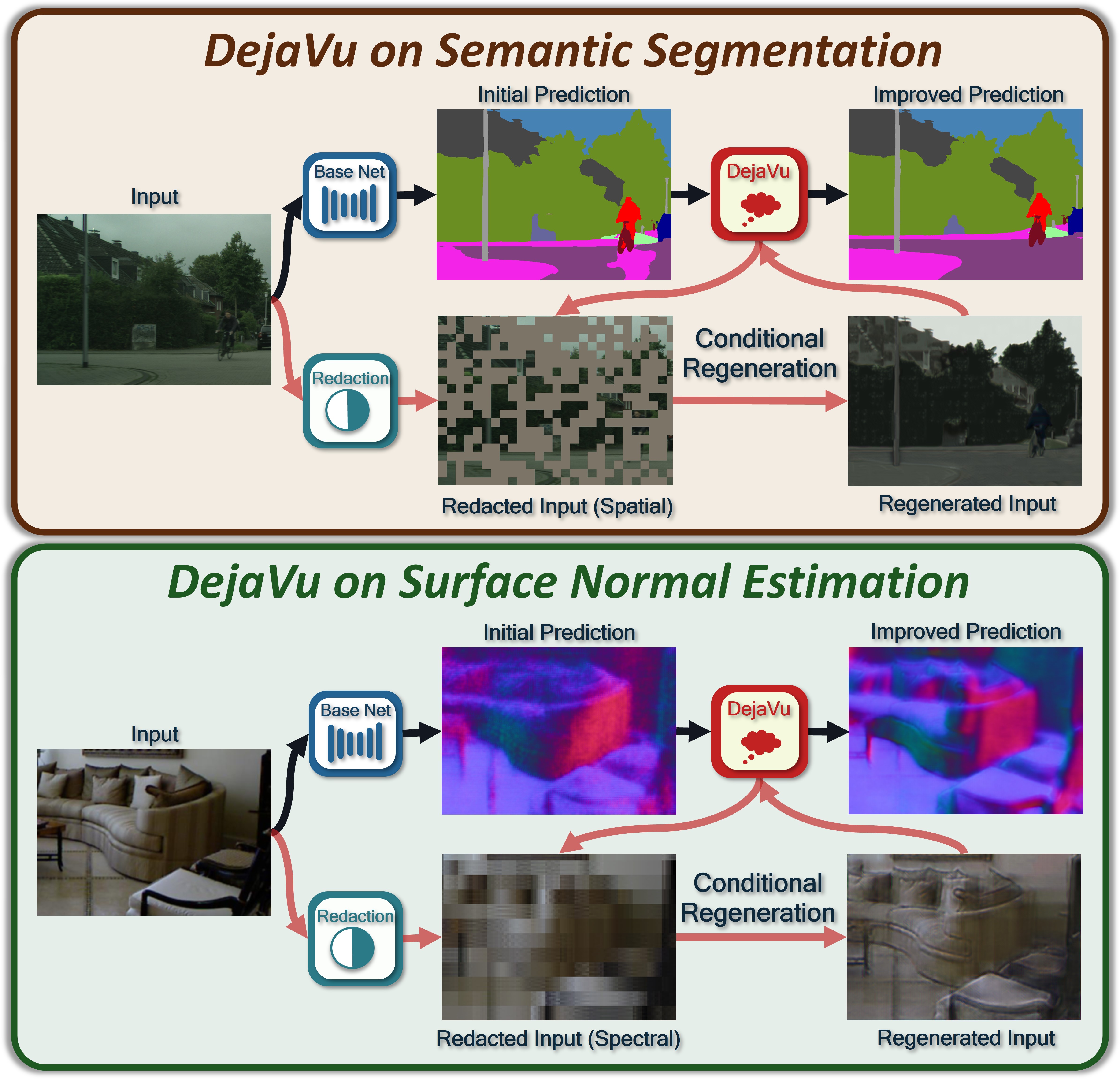}
\end{center}
\vspace{-15pt}
\caption{Training within the DejaVu framework enables dense prediction models to improve their initial predictions using our proposed loss. The segmentation results are for the same OCR~\cite{yuan2020object} model with and without DejaVu. The surface normal results are for SegNet-XTC~\cite{li2022learning}.   
}
\vspace{-10pt}
\label{fig:dejavu}
\end{figure}

Dense prediction tasks produce per-pixel classification or regression results, such as semantic or panoptic class labels, depth or disparity values, and surface normal angles. These tasks are critical for many vision applications to better perceive their surroundings for XR, autonomous driving, robotics, visual surveillance, and so on. There has been significant success in adopting neural networks to solve dense prediction tasks through innovative architectures, data augmentations and training optimizations. For example,~\cite{lin2017focal} addresses pixel level sampling bias and \cite{borse2021inverseform} incorporates boundary alignment objectives. Orthogonal to existing methods, we explore novel regeneration-based ideas to understand how additional gradients from reconstruction tasks complement the established training pipelines for dense prediction tasks and input representations.

There are works \cite{sun2020conditional, zhang2016augmenting} in classification settings that leverage reconstructions and likelihood-based objectives as auxiliary loss functions to enhance the quality of feature representations 
and also improve Open-set/OOD Detection \cite{nalisnick2019hybrid,pmlr-v80-liu18e,liu2022pac,garrepalli2022oracle}.
The  core intuition is that, for discriminative tasks the model needs a minimal set of features to solve the task and any feature which does not have discriminative power for the target subset of data are ignored. 
Another line of work for dense predictions \cite{lu2020depth} focuses on depth completion and leverages reconstruction-based loss to learn complementary image features that aid better capture of object structures and semantically consistent features. Following such intuitions, we can see that reconstruction-based auxiliary loss should capture more information in representation than discriminative-only training. 

Here, we introduce a novel training strategy, DejaVu
\footnote{In training, DejaVu redacts the input image and constructs its regenerated versions, in a way, these regenerated versions are "already seen" yet not exactly the same due to initial redaction.}, for dense prediction tasks with an additional, conditional reconstruction objective to improve the generalization capacity of the task-specific base networks as illustrated in Fig.~\ref{fig:dejavu}. We redact the input image to remove structure information (e.g., boundaries) while retaining contextual information. We adopt various redaction techniques that drop out components in spatial or spectral domains. Then, we enforce a conditional regeneration module (CRM), which takes the redacted image and the base network's dense predictions, to reconstruct the missing information. For regeneration feasibility, the CRM will require structure information from the dense predictions. By including this conditional regeneration objective during training, we encourage the base network to learn and use such structure information, which leads to more accurate predictions with clearer boundaries and better spatial consistency, as shown in the experimental section. In comparison, the supervised loss cannot capture this information alone since the cross-entropy objective (for segmentation, as an example) looks at the probability distribution of every pixel. In this sense, DejaVu can implicitly provide cues to the dense prediction task from the reconstruction objective depending on the type of redaction we select. We also note that using the same number of additional regenerated images as a data augmentation scheme does not provide the performance improvements that DejaVu can achieve (as reported in the Appendix). This shows that DejaVu conditions the training process more effectively than any data augmentation technique. 




Our DejaVu loss can be applied to train any dense prediction network and does not incur extra computation at test time. When it is feasible to leverage additional computation, DejaVu can be extended where we incorporate an attention-based regeneration module within the dense prediction network, further improving accuracy. An advantage of regenerating the original image from predictions is that we can additionally use other losses including text supervision and cyclic consistency, as described in Section~\ref{subsec:attention}.

Our extensive experiments on multiple dense prediction tasks, including semantic segmentation, depth estimation, and surface normal prediction, show that employing DejaVu during training enables our trained models to outperform the latest state of the art on several large-scale benchmarks.

Our main contributions are summarized as follows:
\vspace{-2mm}
\begin{itemize}
    \item We devise a novel learning strategy, DejaVu, that leverages conditional image regeneration from redacted input images to improve the overall performance on dense prediction tasks. (Sec.~\ref{subsec:recon})
\vspace{-2mm}
    \item We propose redacting the input image to enforce the base networks to learn accurate dense predictions such that these tasks can precisely condition the regenerative process. (Sec.~\ref{subsec:redaction})
\vspace{-2mm}
    \item We devise a novel shared attention scheme, DejaVu-SA, by incorporating the regeneration objective into the parameters of the network. (Sec.~\ref{subsec:attention})
\vspace{-2mm}
    \item We further provide extensions to DejaVu, such as the text supervision loss DejaVu-TS and Cyclic consistency loss DejaVu-CL, further improving performance when additional data is available. (Sec.~\ref{subsec:extension})
\vspace{-2mm}
    \item DejaVu is a universal framework that can enhance the performance of multiple networks for essential dense prediction tasks on numerous datasets with no added inference cost. (Sec.~\ref{sec:experiments})
\end{itemize}

\section{Related Work}
\label{sec:related}\vspace{-3pt}

\textbf{Dense Prediction with Supervised Learning}: 
Deep learning has been successfully applied for various dense prediction tasks such as semantic segmentation \cite{long2015fully}, with hierarchical branches \cite{deeplabV2,deeplabV3,zhao2017pspnet}, attention mechanisms \cite{wang2018non,fu2019dual,yuan2018ocnet,huang2019ccnet,borse2023x,strudel2021segmenter}, and auxiliary losses \cite{borse2021hs3,borse2021inverseform}, to point out a few. There are also extensions to panoptic and instance segmentation \cite{xiong19upsnet,kirillov2019panopticfpn,cheng2020panoptic,li2021fully}, plug and play modules \cite{zhang2021knet,borse2022panoptic}, and universal architectures \cite{cheng2021maskformer,cheng2022masked}. To improve performance, many works utilize extra image/text datasets or use test-time augmentation such as multi-scale inference~\cite{wang2022image, su2022towards, wang2022internimage, fang2022eva }.
Depth estimation is another dense prediction task where most works usually adopt self-supervised training paradigms \cite{godard2017unsupervised} in the stereo setting or \cite{guizilini20203d,godard2019digging,zhou2017unsupervised,patil2020don,jiang2020dipe} in monocular videos, exploiting geometric transformation and consistency between views to train the network. Many works leverage semantics \cite{guizilini2019semantically,klingner2020self,Cai2021xdistill} and consistency~\cite{qi2018geonet,qi2020geonet++, zhang2022perceptual}, spherical regression \cite{liao2019spherical}, uncertainty \cite{bae2021estimating} and spatial rectifiers \cite{do2020surface}. Some efforts simultaneously solve these tasks within multi-task settings, exploiting inter-task information \cite{liu2019end,chen2018gradnorm,kendall2017multi,sener2018multi,li2022learning, klingner2023x, zhang2022auxadapt}. In our work, we present a framework and training objective that can enhance the performance of such dense prediction tasks individually or in a multi-task setting.

\textbf{Image Reconstruction as an Auxiliary Task}:
Autoencoders \cite{hinton1993autoencoders} and their variants \cite{vincent2008extracting} are among popular techniques for reconstruction-based unsupervised representation learning. Recent works incorporate different masking \cite{vincent2010stacked,pathak2016context,chen2020generative} and channel removal~\cite{zhang2016colorful} schemes. Inspired by success in NLP~\cite{chen2020generative}, masking-inspired ideas are exploited to learn image representations, for instance, \cite{dosovitskiy2020image, bao2021beit} via image tokens, \cite{he2022masked} in pretraining, \cite{karnam2020self}, \cite{karnam2020self, xia2017w} in self-supervised segmentation, and \cite{di2021pixel} in detection. Reconstruction-based objectives are also used for domain adaptation \cite{yang2020label}. In comparison, our work aims at making the best use of image reconstruction to improve supervised image segmentation. 

\textbf{Image Generation \& Translation:}
Pixel-to-pixel image translation~\cite{liu2019learning, park2019semantic, wang2018high} aims to improve the quality of image synthesis conditioned on segmentation maps. In contrast, we use reconstruction to enhance the dense prediction quality. There are also image stylization methods that blend two images into a new one, keeping the content of the one while changing the style according to the other. \cite{ulyanov2017improved,sanakoyeu2018style,karras2020analyzing}. 
More recently, diffusion-based models have demonstrated remarkable reconstruction performance \cite{rombach2022high, bansal2022cold}. In our work, we also consider an extension motivated by denoising diffusion. However, instead of training within the diffusion paradigm, we design an iterative generator module that reduces the degree of masking across its timesteps using dense predictions and redacted images.

\begin{figure}[t]
\begin{center}
\includegraphics[width=0.99\linewidth]{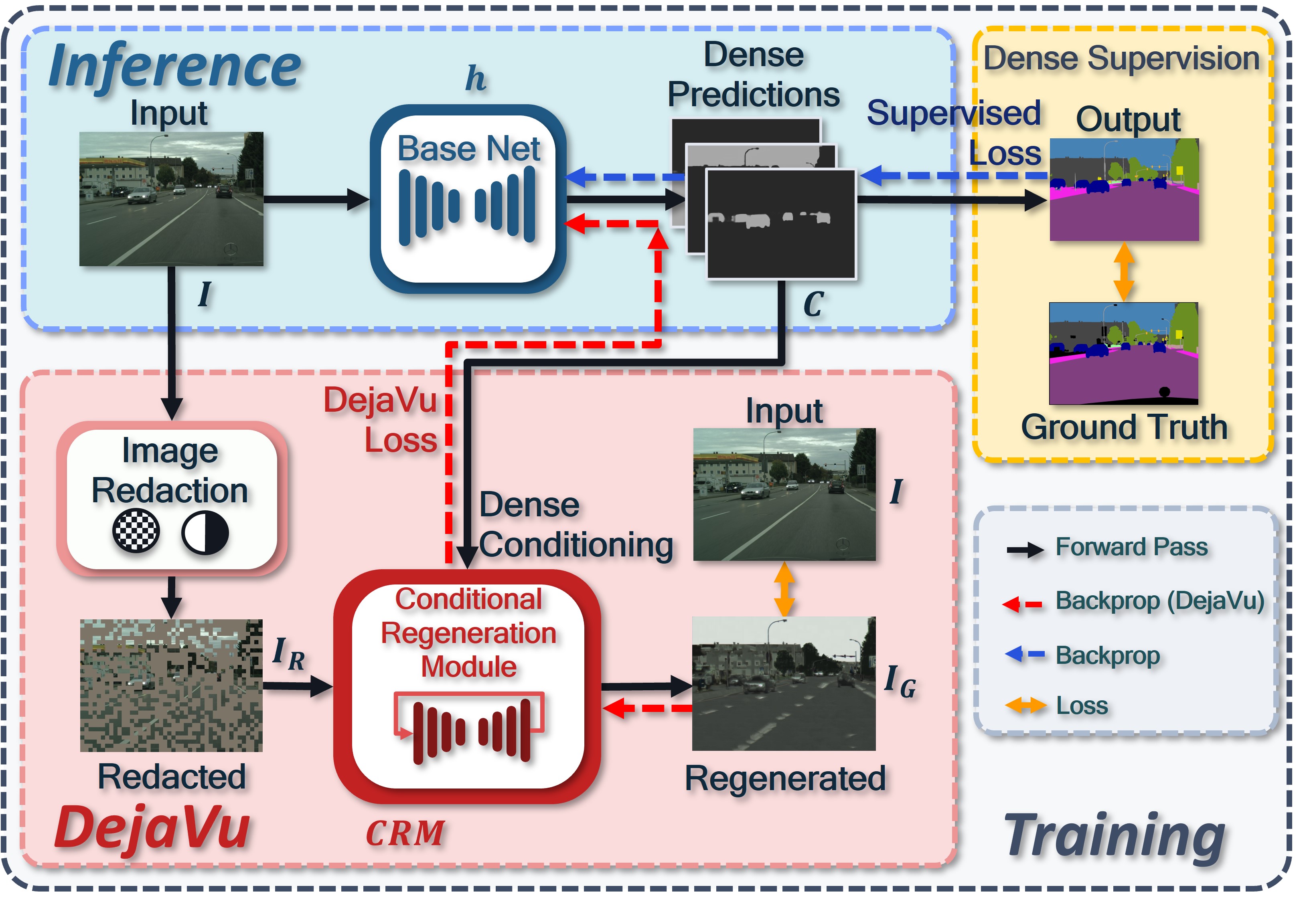}
\end{center}
\vspace{-20pt}
\caption{DejaVu consists of image redaction, dense conditioning from a base network (such as class-wise masks for semantic segmentation), and a conditional regeneration module (CRM) that constructs a regenerated version of the input image, from a redacted image and dense conditioning. The DejaVu loss backpropagates to update the base network with the supervised loss. 
}
\vspace{-10pt}
\label{fig:masked-training}
\end{figure}

\section{DejaVu Framework}
\label{sec:proposal}
\vspace{-3pt}
In this Section, we discuss the DejaVu framework in detail, along with all its extensions. 

\textbf{Overview:} As illustrated in Fig.~\ref{fig:masked-training}, consider that we train a dense prediction network (called the base net) $h$, which inputs an image $I$ to generate dense outputs $C = h(I)$. We first apply redaction to the input $I$, generating the redacted image $I_R$. The redaction methods are explained in Section~\ref{subsec:redaction}. Next, we pass the redacted image $I_R$ and dense outputs $C$ as inputs to a conditional regeneration module (CRM), as described in Section~\ref{subsec:arch}. The CRM outputs a regenerated image $I_G$, which is then compared with the original image to provide a loss for training, as explained in Section~\ref{subsec:recon}. We also present an optional shared attention module, DejaVu-SA, which integrates the regeneration operation from the DejaVu loss into the base net in inference (Section~\ref{subsec:attention}). Finally, we elaborate on further extensions of DejaVu that incorporate vision-language training and consistency regularization (Section~\ref{subsec:extension}).

In the following subsection, we describe image redaction options.

\subsection{Image Redaction }
\label{subsec:redaction}
\vspace{-3pt}

\begin{figure}[t]
\begin{center}
\includegraphics[width=0.96\linewidth]{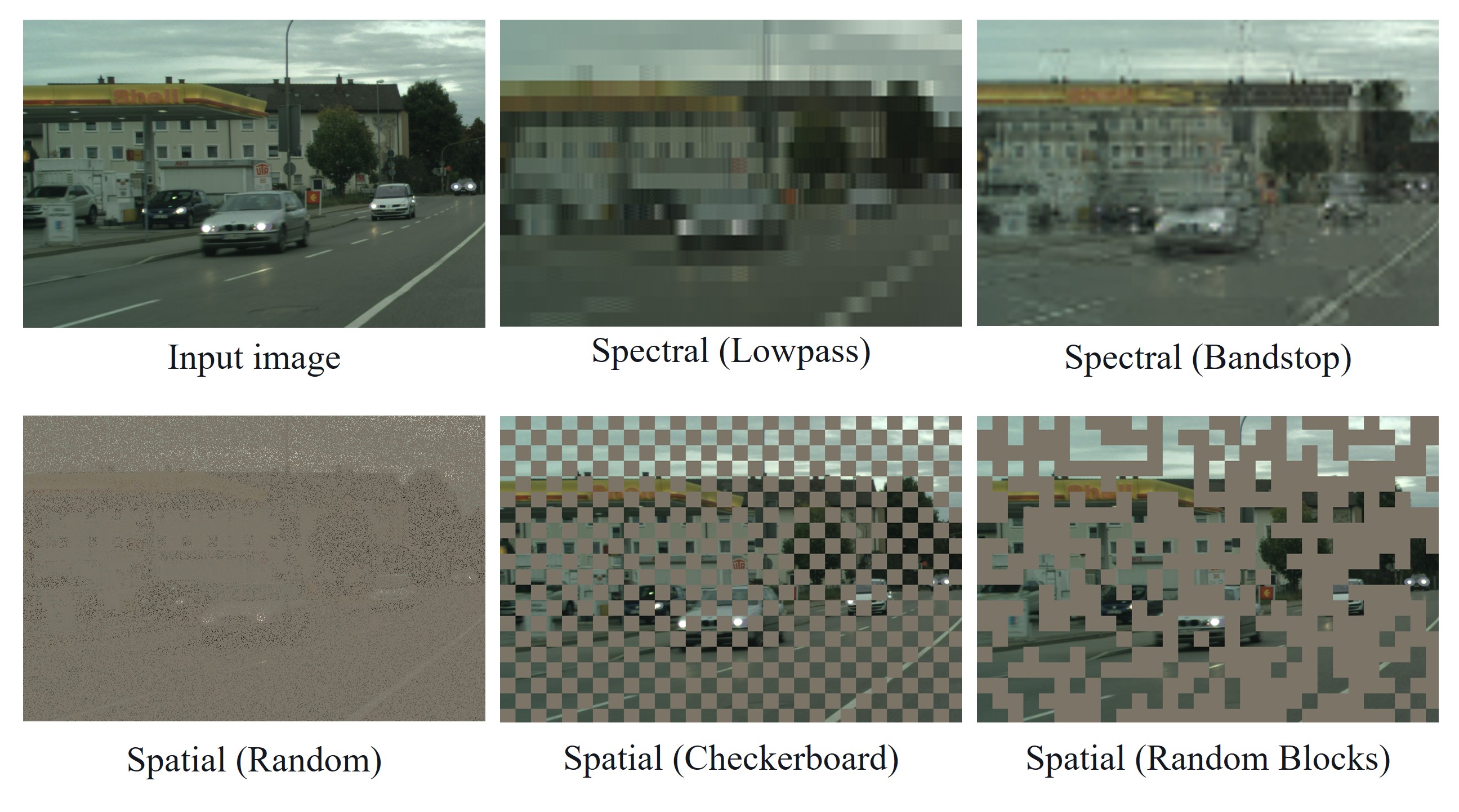}
\end{center}
\vspace{-20pt}
   \caption{Sample redactions in spectral and spatial domains.}
\vspace{-10pt}
\label{fig:redaction}
\end{figure}

When applying redaction to the input image, we intentionally remove the type of information that would be desired for the dense prediction base network to learn to generate. 
Selecting an appropriate redaction style is critical to ensure the image regeneration is feasible and the dense prediction network receives useful feedback to learn better features. We consider information redaction in two domains, i.e., spatial and spectral, which target different image attributes.

When performing spatial redaction, we mask out specific pixels. The original pixel values can be removed randomly or in a structured fashion. Figure~\ref{fig:redaction} shows typical examples of random and structured spatial redaction. 
For instance, in random spatial redaction, we randomly mask pixels with a fixed probability $t$. Alternatively, we generate a checkerboard redaction by setting a block size of $b$. We also extend the checkerboard redaction by randomly shifting the grids, to generate the random blocks redaction. The values of $t$ and $b$ are considered as hyperparameters. As visible, structural motifs and details, such as object silhouettes and semantic class boundaries, are partially removed with spatial redaction. As a result, the CRM will enforce and thus facilitate the dense prediction network to embed such removed and missing information in the predictions for the CRM to regenerate the original image as closely as possible.

As for spectral redaction, we first transform the original image to the frequency domain, e.g., Discrete Cosine Transform (DCT). We mask out DCT components and then apply the inverse transform to obtain the redacted image. As shown in Fig.~\ref{fig:redaction}, lowpass spectral redaction implies masking out high-frequency DCT components of the input image, while highpass spectral redaction implies masking out low-frequency components of the input image. We also experiment with bandstop redactions, by masking out a band from the middle. Filtering out high-frequency coefficients causes finer image patterns to be distorted, and applying bandstop filters smears object-level details at certain scales. When providing such redacted images for regeneration, the dense prediction network will be required to embed the corresponding information (e.g., distorted patterns, edges, object details) in its predictions.

Based on our analysis shown in Section~\ref{sec:experiments}, we empirically observe that spatial redaction performs better on segmentation tasks as the DejaVu loss penalizes inaccurate class-wise predictions (similar to the one observed in Fig.~\ref{fig:dejavu}) heavily. On the other hand, spectral redaction works well on depth and surface normal estimation tasks as their outputs are penalized heavily if they contain blurred boundaries. 

In the following subsection, we describe the Conditional Regeneration Module architecture.

\subsection{Conditional Regeneration Module}
\label{subsec:arch}\vspace{-3pt}

\begin{figure}[t]
\begin{center}
\includegraphics[width=0.98\linewidth]{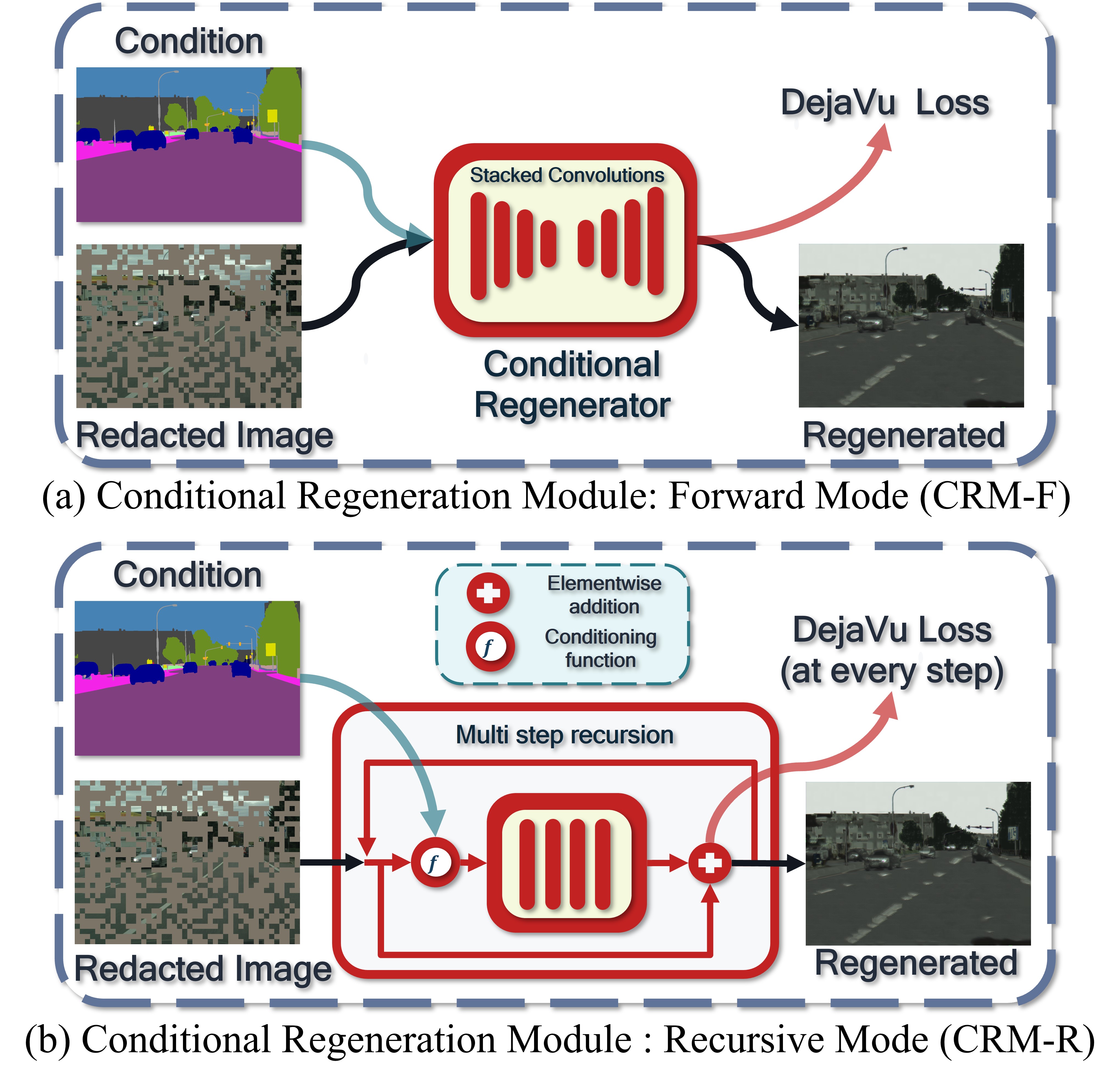}
\end{center}
\vspace{-20pt}
\caption{Architecture of the CRM. Regardless of the dense prediction task, each mode can cope with redactions effectively, the Forward (one-step) mode for structured redactions and the Recursive mode for random redactions.} 
\vspace{-10pt}
\label{fig:conditional}
\end{figure}

The Conditional Regeneration Module (CRM) takes the redacted image $I_R$ as well as the dense conditioning $C$, and regenerates the input image $I$. 
We use two types of regeneration modes: 1) Forward mode CRM-F, and 2) Recursive mode CRM-R, as shown in Fig.~\ref{fig:conditional}.
Both modules take the dense condition and the redacted image as an input and produce perceptually similar images to the original input image at the output.
CRM-F, illustrated in Fig.~\ref{fig:conditional}(a), simply consists of stacked Conv-BatchNorm-ReLU blocks, inspired by the related work on conditional image translation~\cite{park2019semantic}.

As the CRM-F performs an auxiliary reconstruction task, there exists a trade-off between its model complexity and the dense prediction task accuracy. We study this trade-off in the Appendix.
CRM-R, illustrated in Fig.~\ref{fig:conditional}(b), consists of a single Convolution-BatchNorm-ReLU block, which recursively produces a residual for the reconstructed image, inspired by cold diffusion~\cite{bansal2022cold}. The number of steps is treated as a hyperparameter over which we perform a search. Hyperparameter choices for both CRM architectures are provided in the Appendix. 
Empirically, we find that the CRM-R is more effective for random occlusions and CRM-F produces better results on structured occlusions.

Inputs to the CRM are the dense condition $C\in \mathbb{R}^{N\times H\times W}$ and redacted input image $I_R\in \mathbb{R}^{3\times H\times W}$, where $H$ and $W$ are height and width, and $N$ is the number of predicted channels. We use two operations to combine the two inputs: 1) multiplication and 2) concatenation. 
For the multiplication operation, we average the input image to a single channel, $\overline{I_R}\in \mathbb{R}^{1\times H\times W}$, broadcast to $N$ channels, and perform element-wise channel multiplication with the dense condition.
Likewise for the concatenation operation, $I_R$ and $C$ are concatenated along the channel dimension and fed as input to the generation module, and thus the size of input channels is $3 + N$. 

\subsection{DejaVu Loss to Update Base Network}
\label{subsec:recon} \vspace{-3pt}

The DejaVu loss, as illustrated in Fig.~\ref{fig:masked-training}, is computed by comparing the regenerated image $I_G$ to the input image $I$. We add this loss to the original task loss term during training. Specifically, when comparing the original and regenerated images, we use the Mean Squared Error (MSE) and LPIPS losses~\cite{zhang2018unreasonable} in order to attain supervision on both local (pixel-level) and global (content-level) feedback. The total training loss is given as follows: \vspace{-5pt}
\begin{equation} \label{eq:loss1}
     \mathcal{L} = \mathcal{L}_\text{base} + \mathcal {\gamma}\cdot \mathcal{L}_\text{regen}, \vspace{-5pt}
\end{equation}
\begin{equation}
    \label{eq:loss2}
     \mathcal{L}_\text{regen} = \mathcal {\gamma}_1 \cdot \mathcal{L}_\text{lpips} + {\gamma}_2\cdot \mathcal{L}_\text{mse},
\end{equation}
where $\mathcal{L}_\text{base}$ is the loss from the base training procedure, e.g., cross-entropy loss for semantic segmentation, $\mathcal{L}_1$ loss for depth estimation.
$\mathcal{L}_\text{regen}$ is the loss from our proposed conditional image regeneration. $\gamma$, $\gamma_1$, and $\gamma_2$ are hyperparameters that blend these loss terms.

\subsection{DejaVu Shared Attention Module (DejaVu-SA)}
\label{subsec:attention}\vspace{-3pt}

\begin{figure}[t]
\begin{center}
\includegraphics[width=0.98\linewidth]{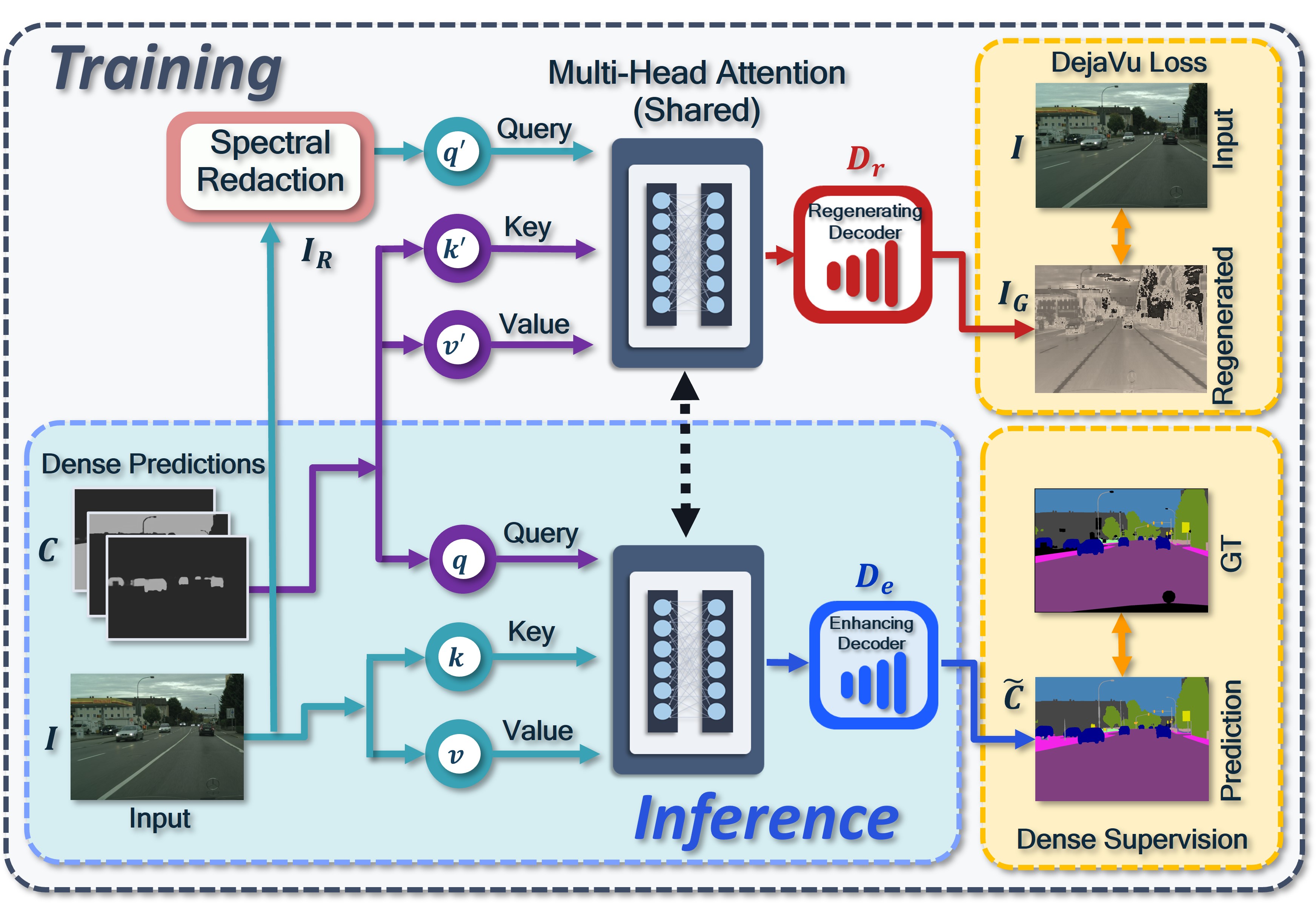}
\end{center}
\vspace{-10pt}
\caption{Folding the CRM into a shared attention mechanism. For dense supervision, a multi-head attention block consumes dense predictions as queries while inputs as keys \& values. For regeneration, the same block consumes spectrally redacted inputs as queries and dense predictions as keys \& values.}
\label{fig:dejavuattn}
\vspace{-10pt}
\end{figure}



The DejaVu loss provides improvements at no added inference cost. However, in cases when we have room for increasing the computational complexity of our base network, we propose the DejaVu Shared Attention scheme, DejaVu-SA, which incorporates the DejaVu operation into the parameters of our base network. 
The inputs to DejaVu-SA are the image $I$ and dense predictions $C$, and the outputs are enhanced predictions $\tilde{C}$. The attention module is illustrated in Fig.~\ref{fig:dejavuattn}. 

During training, we perform two passes through the attention module. We first push the input image $I$ through the spectral redaction operation to obtain the redacted input $I_R$.
We compute queries $Q^{s}_{I_R} = q'(I_R)$, keys $K^{s}_C = k'(C)$ and values $V^{s}_C = v'(C)$, where $q'$, $k'$ and $v'$ denote patch embedding operations in Fig.~\ref{fig:dejavuattn}. Through a multi-head attention (MHA) operation followed by rearrangement, we pass the outputs through regeneration decoder $D_r$ to obtain the regenerated image $I_G = D_r(\text{MHA}(Q^{s}_{I_R}, K^{s}_C, V^{s}_C))$. This regenerated image is now supervised by with the original image $I$.
In the second pass, we obtain queries $Q^{s}_C  = q(C)$, keys $K^{s}_I = k(I)$ and values $V^{s}_I = v(I)$ where $q$, $k$ and $v$ denote patch embedding operations shown in Fig.~\ref{fig:dejavuattn}. Through the same MHA operation followed by rearrangement, we pass the outputs through enhancement decoder $D_e$, to obtain enhanced predictions $\tilde{C} = D_e(\text{MHA}(Q^{s}_C, K^{s}_I, V^{s}_I))$. The enhanced predictions $\tilde{C}$ are used as the final predictions and are supervised by ground truth for training the whole network. The intermediate channel dimension after patch embedding and embedding dimension in the MHA module are treated as constants and used as a hyperparameter. In Fig.~\ref{fig:dvsa_complexity}, we provide results for the trade-off between accuracy and complexity, after scaling these intermediate dimensions.

\subsection{Extending the Dejavu framework}
\label{subsec:extension}\vspace{-3pt}
In this section, we describe extensions of the DejaVu framework that can produce further enhancements to dense predictions after regenerating the input image.

\textbf{Regenerated Text Supervision (DejaVu-TS):}
After regenerating the input image from the dense prediction, we propose to perform a novel text-based supervision objective. More specifically, we match CLIP~\cite{radford2021learning} features between the original image $I$ and the regenerated image $I_G$. Essentially, we can obtain the CLIP features $f_G = \text{CLIP}(I_G)$ and $f_I = \text{CLIP}(I)$ for the reconstructed image and the input image respectively. Optionally, the CLIP model can be conditioned by only the tokenized input of the class names for segmentation tasks. The matching loss is the mean squared error between the features and can be defined as
\begin{equation}\label{eq:text}
\mathcal{L}_{text} = (1/D)||f_G - f_I||^{2}_2
\end{equation}
where $D$ is the number of elements in the feature vectors.

\textbf{Cyclic Consistency loss (DejaVu-CL):}
Another benefit of regenerating the input image is our proposed Cyclic Consistency Loss. Once the regenerated image $I_G$ is produced, we propose to pass it through the base network $h$ to produce the regenerated predictions $C_G$. We apply the MSE loss to match the outputs $C_G$ with the dense predictions $C$. The detailed structure and results for this loss term are illustrated in the Appendix.

\begin{figure*}[t]
\begin{center}
\includegraphics[width=0.96\linewidth]{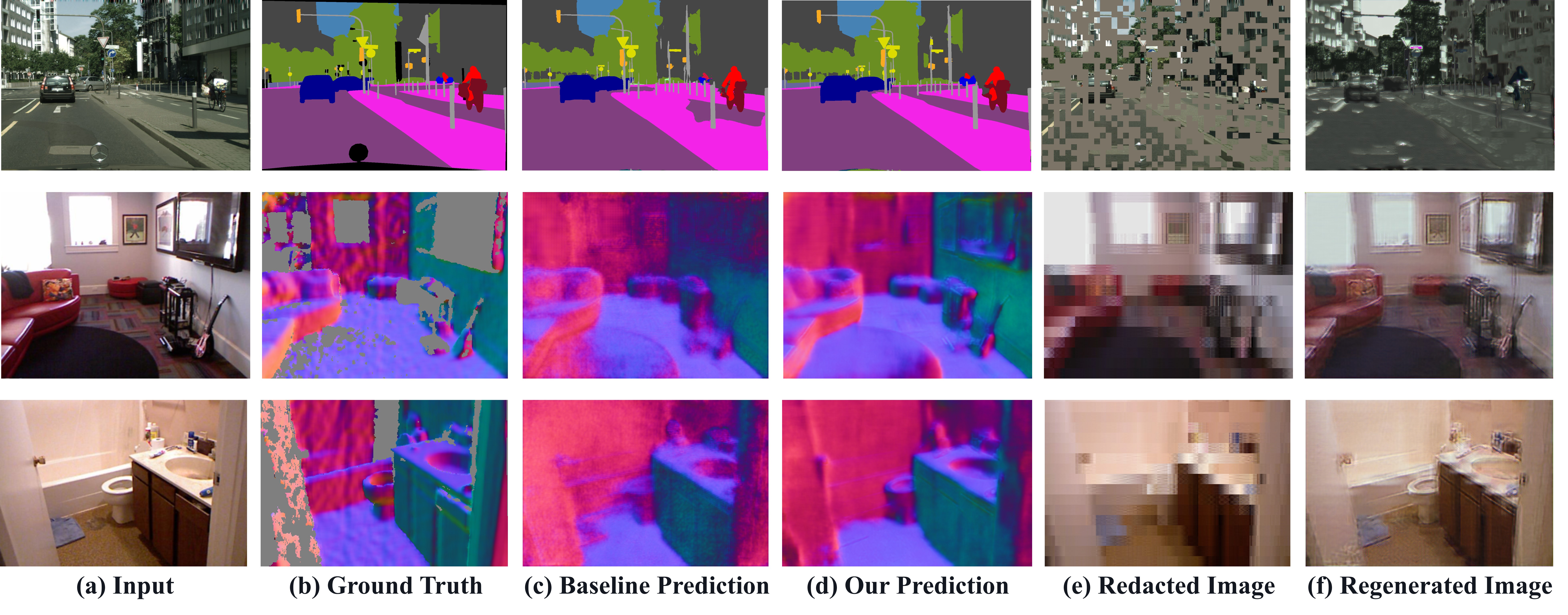}
\end{center}
\vspace{-7mm}
\caption{Visualization of (c) baseline prediction, (d) our enhanced prediction and the (f) regenerated image given the (a) input image, (b) ground truth and (e) redacted input. This is shown for three dense prediction tasks: semantic segmentation on Cityscapes using OCR (top) and surface normal estimation on NYUD-v2 using SegNet-XTC (middle and bottom).}
    \vspace{-8pt}
    \label{fig:qual}
\end{figure*}

\section{Experimental Setup}
\label{sec:experiments}\vspace{-3pt}
In this section (and the Appendix), we perform comprehensive analysis and experiments using DejaVu. 
\subsection{Implentation Details}\vspace{-3pt}
\textbf{Datasets:} For Semantic Segmentation we consider Cityscapes~\cite{cordts2016cityscapes}, and ADE20K~\cite{zhou2017scene} datasets. 
Cityscapes consists of 5,000 annotated images of size $1024\! \times \! 2048$, divided into 2,975 training, 500 validation, and 1,525 test images. It covers 11 stuff and 8 thing classes.  
For Panoptic Segmentation we consider COCO~\cite{lin2014microsoft}, COCO consists of 118,000 training, 5,000 validation, and 20,000 test images. There are 53 stuff and 80 thing classes in COCO. ADE20K contains 20,210 training, 2,000 validation, and 3,000 test images, with 35 stuff and 115 thing classes. For Monocular Depth Estimation we consider KITTI\cite{geiger2013vision} and use data split from \cite{eigen2014depth}. 
For the multi-task learning setup, We also consider NYU-Depth-v2 \cite{silberman2012nyu} dataset which has annotations/ground truth for semantic segmentation, Depth and surface-normals. We use the original 795 train and 654 test images for the NYUD-v2 dataset.

\textbf{Networks and Training:}
We train multiple existing baselines using the DejaVu loss term for various dense prediction tasks. For semantic segmentation with Cityscapes dataset, we apply DejaVu to HRNet~\cite{WangSCJDZLMTWLX19hrnet}, OCR~\cite{wang2019object} and HMS~\cite{tao2020hierarchical}. For semantic segmentation with ADE20K dataset, we apply DejaVu loss to Semantic FPN~\cite{kirillov2019panoptic}, UperNet~\cite{xiao2018unified} and DenseCLIP~\cite{rao2022denseclip}. Semantic FPN uses PoolFormer~\cite{yu2022metaformer} backbone while the rest use ViT backbone~\cite{dosovitskiy2020image}. For panoptic segmentation, we train Mask2Former~\cite{cheng2022masked} with Swin~\cite{liu2021swin} backbones using the DejaVu loss. For depth estimation, we apply DejaVu loss to self-supervised training of Monodepth2~\cite{godard2019digging}. For fully and partially supervised multi-task learning setting, we apply DejaVu loss to MTL~\cite{caruana1997multitask} and XTC~\cite{li2022learning} baseline models. For all base methods, we apply the original loss function ie. cross-entropy loss for semantic segmentation, $\mathcal{L}_1$ norm loss for depth estimation, cosine similarity loss
for surface normal estimation, etc. in addition to the Déjà vu loss described in Eq.~\eqref{eq:loss1}.
Training details are all discussed in the Appendix.


\textbf{Evaluation Metrics:}
For semantic segmentation tasks, we evaluate using mean intersection-over-union mIoU. For the panoptic segmentation task, we report panoptic quality PQ~\cite{kirillov2019panoptic}. We also report PQ scores for things and stuff, denoted as PQ$^{th}$ and PQ$^{st}$, respectively. In the multi-task learning setup~\cite{liu2019end}, depth estimation performance is evaluated using absolute relative error aErr, and surface normal estimation performance is evaluated using mean error mErr in the predicted angles. For monocular depth estimation, we use absolute relative error Abs Rel and squared relative error Sq Rel. Furthermore, the classification metric $\delta_1$ measures whether the ratio between ground truth and estimated depth values is within a certain range around 1. We also report GMacs to measure efficiency.


\begin{table}[h]
\captionsetup{font=small, belowskip=-10pt}
\small
\centering
\setlength{\tabcolsep}{3.2pt}
    \centering
    \begin{tabular}{c|l|ccc}
    \toprule
    \textit{Backbone} & 
    \textit{Method} & 
    \cellcolor{BurntOrange}\textit{mIoU}$\uparrow$ & \cellcolor{Rhodamine}\textit{GMacs}$\downarrow$\\
    \midrule
    \multirow{6}{*}{HRNet18} & HRNet~\cite{WangSCJDZLMTWLX19hrnet} & 77.6 & 19 \\
    &\cellcolor{gray85}\textbf{+DejaVu} &\cellcolor{gray85}78.8 &\cellcolor{gray85}19\\
    & HS3\cite{borse2021hs3} & 78.1 & 19 \\
    & HS3-Fuse\cite{borse2021hs3} & 81.4 & 39\\
    & OCR~\cite{wang2019object} & 80.7 & 39\\
    &\cellcolor{gray85}\textbf{+DejaVu} &\cellcolor{gray85} \textbf{82.0} & \cellcolor{gray85}39\\
    \midrule
    MiT-B5 & Segformer~\cite{xie2021segformer} & 84.0 & 362\\
    \midrule
    \multirow{2}{*}{Swin-L~\cite{liu2021swin}} & Mask2Former~\cite{cheng2022masked} & 83.3 & 251\\
    & SeMask~\cite{jain2021semask} & 84.0 & 258\\
    \midrule
    ViT & ViT Adapter~\cite{chen2022vitadapter} & 84.9 & 1089\\
    \midrule
    \multirow{6}{*}{HRNet48} & HRNet & 84.7 & 175\\
    &\cellcolor{gray85}\textbf{+DejaVu} & \cellcolor{gray85}85.4 & \cellcolor{gray85}175\\
    & OCR & 86.1 & 348\\
    &\cellcolor{gray85}\textbf{+DejaVu} & \cellcolor{gray85}86.5 & \cellcolor{gray85}348\\
    & HMS~\cite{tao2020hierarchical} & 86.7 & 893\\
    &\cellcolor{gray85}\textbf{+DejaVu} & \cellcolor{gray85}\textbf{87.1} & \cellcolor{gray85}893\\
    \bottomrule
    \end{tabular}
    \vspace{-5pt}
    \caption{Comparing with SOTA methods on Cityscapes val.}
    \label{tab:cityval-main}
    \vspace{-5pt}
\end{table}

\subsection{Experiments using the DejaVu Loss}\vspace{-3pt}

In this subsection, we perform experiments by training various baseline models using our proposed DejaVu loss function.

\textbf{Semantic Segmentation:} In Table~\ref{tab:cityval-main}, we report results of comparing our proposed framework with respect to semantic segmentation baselines on the Cityscapes val dataset. As observed, our method can produce boost of more than \textbf{1.3} pts in mIoU when using HRNet18 backbones. When using heavier HRNet48 versions, training with the DejaVu loss still improves, but the relative improvement reduces closer to SOTA scores. We also train HRNet-OCR~\cite{yuan2020object} and HRNet-OCR-HMS~\cite{tao2020hierarchical} backbones using the DejaVu loss. The HMS backbone trained with DejaVu loss acheives the SOTA score on Cityscapes val. In Table~\ref{tab:adeval}, we show improvement over existing semantic segmentation baselines on the ADE20K val dataset. Specifically, adding DejaVu loss on top of Semantic FPN, UPerNet and DenseCLIP produces consistent improvements in mIoU. 

\begin{table}[h]
\captionsetup{font=small, belowskip=-10pt}
\small
\centering
\setlength{\tabcolsep}{3.2pt}
    \centering
    \begin{tabular}{l|l|ccc}
    \toprule
    \textit{Method} & 
    \textit{Backbone} & 
    \cellcolor{Lavender}\textit{PQ}$\uparrow$ & \cellcolor{Salmon}\textit{PQ$^{st}$}$\uparrow$ &
    \cellcolor{Goldenrod}\textit{PQ$^{th}$}$\uparrow$ \\
    \midrule
    MaX-Deeplab~\cite{wang2021max} & Max-S & 48.4 & 53.0 & 41.5 \\
    MaskFormer~\cite{cheng2021maskformer} & Swin-T & 47.7 & 51.7 & 41.7 \\
    Mask2Former~\cite{cheng2022masked} & Swin-T & 53.2 & 59.3 & 44.0 \\
    \cellcolor{gray85}\textbf{+DejaVu} &\cellcolor{gray85} Swin-T & \cellcolor{gray85}\textbf{54.3} & \cellcolor{gray85}\textbf{60.5} & \cellcolor{gray85}\textbf{44.9} \\
    \midrule
    MaX-Deeplab~\cite{wang2021max} & Max-L & 51.1 & 57.0 & 42.2 \\
    K-Net~\cite{zhang2021knet} & Swin-L & 54.6  & 60.2  & 46.0 \\
    MaskFormer~\cite{cheng2021maskformer} & Swin-L &  52.7 & 58.5 & 44.0 \\
    Mask2Former~\cite{cheng2022masked} & Swin-L & 57.6 & 64.2 & 47.5 \\
    \cellcolor{gray85}\textbf{+DejaVu} &\cellcolor{gray85} Swin-L &\cellcolor{gray85} \textbf{58.0} &\cellcolor{gray85} \textbf{64.4} &\cellcolor{gray85} \textbf{48.3} \\
    \bottomrule
    \end{tabular}
    \vspace{-5pt}
    \caption{Comparison with SOTA methods on COCO Panoptic Segmentation val.}
    \label{tab:cocoval}
    \vspace{3pt}
\end{table}

\textbf{Panoptic Segmentation:} In Table~\ref{tab:cocoval}, we show results for panoptic segmentation on the COCO val dataset. We trained the previous SOTA method, Mask2former~\cite{cheng2022masked}, using our DejaVu loss function and observed consistent improvements in PQ on two different Swin backbones.

\begin{table}[h]
\captionsetup{font=small, belowskip=-5pt}
\small
\centering
\setlength{\tabcolsep}{3.2pt}
    \centering
    \begin{tabular}{l|c|ccc}
    \toprule
    \textit{Method} & 
    \textit{Backbone} & 
    \cellcolor{BurntOrange}\textit{mIoU}$\uparrow$ \\
    \midrule
    Semantic FPN~\cite{kirillov2019panoptic} & PoolFormer-M48 & 42.4  \\
    \cellcolor{gray85}\textbf{+DejaVu} & \cellcolor{gray85} PoolFormer-M48 & \cellcolor{gray85} \textbf{43.3} \\
    \midrule
    UperNet~\cite{xiao2018unified} & ViT-B~\cite{dosovitskiy2020image}  & 47.4 \\
    \cellcolor{gray85}\textbf{+DejaVu} & \cellcolor{gray85} ViT-B & \cellcolor{gray85} 48.2  \\
    SETR-MLA-DeiT~\cite{zheng2021rethinking} & ViT-B  & 46.2 \\
    Semantic FPN~\cite{kirillov2019panoptic} & ViT-B  & 48.3 \\
    DenseCLIP~\cite{rao2022denseclip} & ViT-B  & 49.8 \\
    \cellcolor{gray85} \textbf{+DejaVu} & \cellcolor{gray85} ViT-B &  \cellcolor{gray85} \textbf{50.3} \\
    \midrule
    Mask2Former~\cite{cheng2022masked} & Swin-L & 56.0  \\
    \cellcolor{gray85}\textbf{+DejaVu} & \cellcolor{gray85} Swin-L & \cellcolor{gray85} \textbf{56.5} \\
    \bottomrule
    \end{tabular}
    \vspace{-5pt}
    \caption{Comparison on ADE20K Semantic Segmentation val.}
    \label{tab:adeval}
\end{table}

\textbf{Self-Supervised Monocular Depth Estimation:} In Table~\ref{tab:monoD}, we show comparison results on monocular depth estimation after training the Monodepth2-R50 baseline using the DejaVu loss. Results show improved performance in terms of lower Abs Rel and Sq Rel error and higher $\delta_1$ compared to other competitive baselines. Interestingly, Monodepth2 produces much higher Sq Rel error compared to the highly competitive PackNet~\cite{guizilini20203d}. However, training with DejaVu reduces the Sq Rel error by a large margin, successfully outperforming PackNet. This shows that DejaVu can also work well with fully self-supervised training schemes, in addition to conventional supervised training.

\begin{table}[h]
\captionsetup{font=small, belowskip=-10pt}
\small
\centering
\begin{tabular}{l|ccc}
    \toprule
\textit{Method}     & 
\cellcolor{Apricot}\textit{Abs Rel}$\downarrow$  & 
\cellcolor{SpringGreen}\textit{Sq Rel }$\downarrow$ &
\cellcolor{SkyBlue}\textit{$\delta_1$}$\uparrow$ \\
\midrule
PackNet-SfM~\cite{guizilini20203d}          & 0.111   & 0.785  & 0.878 \\
Guizilini~\cite{guizilini2019semantically}  & 0.113   & 0.831  & 0.878 \\
Klingner~\cite{klingner2020self}            & 0.112   & 0.833  & 0.884 \\
DiPE~\cite{jiang2020dipe}                   & 0.112   & 0.875  & 0.880  \\
Patil~\cite{patil2020don}                   & 0.111   & 0.821  & 0.883 \\
Monodepth2~\cite{godard2019digging}         & 0.110   & 0.903  & 0.883 \\
\rowcolor{gray85} \textbf{+DejaVu}                            & \textbf{0.108}   & \textbf{0.769} & \textbf{0.885} \\
\bottomrule
    
\end{tabular}\vspace{-5pt}
    \caption{Comparison after training with DejaVu loss on the KITTI eigen split for self-supervised monocular depth estimation.}
    \label{tab:monoD}
    \vspace{5pt}
\end{table}

\textbf{Multi-Task Learning:} We also report results after training multi-task learning models using DejaVu loss, in Tables~\ref{tab:mtpsl_nyud} and~\ref{tab:mtl_nyud}. We add the DejaVu loss to the training objectives of SegNet-MTL and SegNet-XTC baselines provided in~\cite{li2022learning}. Adding the DejaVu loss shows increased semantic segmentation accuracy (mIoU) and decreased depth estimation and surface normal estimation errors in all dense tasks for both the fully supervised setting in Table~\ref{tab:mtl_nyud} and partially supervised setting in Table~\ref{tab:mtpsl_nyud}. 

\begin{table}[t!]
\captionsetup{font=small, belowskip=-10pt}
\small
\centering
\setlength{\tabcolsep}{3.2pt}
    \centering
        \resizebox{0.40\textwidth}{!}{  
    \begin{tabular}{l|l|ccc}
    \toprule
    \textit{Labels} & \textit{Method} & 
    \cellcolor{BurntOrange}\textit{Seg.(mIoU)}$\uparrow$ & \cellcolor{YellowGreen}\textit{Depth(aErr)}$\downarrow$ &
    \cellcolor{Cerulean}\textit{Norm(mErr)}$\downarrow$ \\
    \midrule
    \multirow{4}{*}{Random} & MTL~\cite{caruana1997multitask} & 28.30 & 0.6488 & 32.89 \\
    & \cellcolor{gray85} \textbf{+DejaVu} & \cellcolor{gray85} 30.13 & \cellcolor{gray85} 0.6072 & \cellcolor{gray85} 31.97 \\
    & XTC~\cite{li2022learning} & 34.26 & 0.5787 & 31.06 \\
    & \cellcolor{gray85} \textbf{+DejaVu} & \cellcolor{gray85} \textbf{35.72} & \cellcolor{gray85} \textbf{0.5665} & \cellcolor{gray85} \textbf{29.82} \\
    \midrule
    \multirow{4}{*}{Single} & MTL~\cite{caruana1997multitask} & 26.32 & 0.6482 & 33.31 \\
    & \cellcolor{gray85} \textbf{+DejaVu} & \cellcolor{gray85} 28.07 & \cellcolor{gray85} 0.6264 & \cellcolor{gray85} \textbf{32.02} \\
    & XTC~\cite{li2022learning} & 30.36 & 0.6088 & 32.08 \\
    & \cellcolor{gray85} \textbf{+DejaVu} & \cellcolor{gray85} \textbf{31.02} & \cellcolor{gray85} \textbf{0.5959} & \cellcolor{gray85} 32.15 \\
    \bottomrule
    \end{tabular}}
    \vspace{-5pt}
    \caption{Comparison after training with DejaVu loss on NYUD-v2 for Partially Supervised Multi-Task Learning.}
    \label{tab:mtpsl_nyud}
    \vspace{5pt}
\end{table}

\begin{table}[t!]
\captionsetup{font=small, belowskip=-10pt}
\small
\centering
\setlength{\tabcolsep}{3.2pt}
    \centering
    \resizebox{0.40\textwidth}{!}{  
    \begin{tabular}{l|ccc}
    \toprule
    \textit{Method} & 
    \cellcolor{BurntOrange}\textit{Seg.(mIoU)}$\uparrow$ & \cellcolor{YellowGreen}\textit{Depth(aErr)}$\downarrow$ &
    \cellcolor{Cerulean}\textit{Norm(mErr)}$\downarrow$ \\
    \midrule
    MTL~\cite{caruana1997multitask} & 36.95 & 0.5510 & 29.51 \\
    \rowcolor{gray85} \textbf{+DejaVu} & \textbf{37.40} & \textbf{0.5426} & 28.74 \\
    DWA~\cite{liu2019end} & 36.46 & 0.5429 & 29.45 \\
    GradNorm~\cite{chen2018gradnorm} & 37.19 & 0.5775 & \textbf{28.51} \\
    \midrule
    MTAN~\cite{liu2019end} & 39.39 & 0.5696 & 28.89 \\
    MGDA~\cite{sener2018multi} & 38.65 & 0.5572 & 28.89 \\
    XTC~\cite{li2022learning} & 41.00 & 0.5148 & 28.58 \\
   \rowcolor{gray85} \textbf{+DejaVu} & \textbf{42.69} & \textbf{0.4996} & \textbf{27.49} \\
    \bottomrule
    \end{tabular}}
    \vspace{-5pt}
    \caption{Comparison after training with DejaVu loss on NYUD-v2 for Fully Supervised Multi-Task Learning.}
    \label{tab:mtl_nyud}
\end{table}

\subsection{Ablation Studies and Analyses}\vspace{-3pt}
\textbf{Visualization:} Figure~\ref{fig:qual} shows qualitative results with and without our DejaVu loss. Each row corresponds to different images and tasks, for separate baselines trained with and without DejaVu loss. The first row shows semantic segmentation on Cityscapes. Column (c) shows the baseline prediction without our DejaVu loss while column (d) shows the prediction with the DejaVu loss. We observe that our proposed framework produces better quality semantic and panoptic segmentation masks as it can better perceive the structure of the pavement. Furthermore, the regenerated image (f) structurally resembles the input image (a) with some error margin. In the second and third rows, we report visual results for surface normal estimation on NYUD-v2 where we obtain better quality predictions (d) compared to the baseline (c). Also, the regeneration module has sharpened the spectrally redacted input image (e) to produce the regenerated image (f).

\textbf{Varying the types of Redaction:} In Table~\ref{tab:space_vs_freq_nyu}, we study the effect of applying the different types of redaction as explained in Section~\ref{subsec:redaction}, to various dense tasks. Results are shown on the NYUD-v2 dataset for the single-task learning setup. We pick the "Random Blocks" spatial redaction and apply a bandstop spectral redaction selected for various tasks. For all tasks, redaction improves performance. However, we observe that the semantic segmentation performance has better improvement for the spatial redaction technique while depth and normals perform better when clubbed with spectral redaction. 
We conclude that this is because segmentation is a pixel-wise classification task and hence best complements a spatial in-painting task for regeneration. On the other hand, depth and normal estimation is a regression task and requires texture information (as described by its corresponding spectrum) such that it learns accurate shapes to produce a sharp regeneration. 

\begin{table}[htbp]
\captionsetup{font=small, belowskip=-5pt}
\small
\centering
\setlength{\tabcolsep}{3pt}
    \centering
    \begin{tabular}{l|cc|c}
    \toprule
    \textit{Baseline} & \textbf{DejaVu} & \textbf{DejaVu-TS} &
    \cellcolor{BurntOrange}\textit{mIoU}$\uparrow$ \\
    \midrule
    \multirow{3}{*}{HRNet18} & \xm & \xm & 77.6 \\
    & \cellcolor{gray85} \ch & \cellcolor{gray85} \xm & \cellcolor{gray85}\textbf{78.8} \\
    & \cellcolor{gray85} \ch & \cellcolor{gray85} \ch & \cellcolor{gray85}\textbf{79.2} \\
    \midrule
    \multirow{3}{*}{HRNet18-OCR} & \xm & \xm & 80.7 \\
    & \cellcolor{gray85} \ch & \cellcolor{gray85} \xm & \cellcolor{gray85}\textbf{82.0} \\
    & \cellcolor{gray85} \ch & \cellcolor{gray85} \ch & \cellcolor{gray85}\textbf{82.3} \\
    \bottomrule
    \end{tabular}
    \vspace{-5pt}
    \caption{Effect of applying CLIP supervision (extra data) over the regenerated images to train various baseline models on Cityscapes.}
    \label{tab:DVTS}
\end{table}

\begin{figure}[t]
\begin{center}
\includegraphics[width=0.8\linewidth]{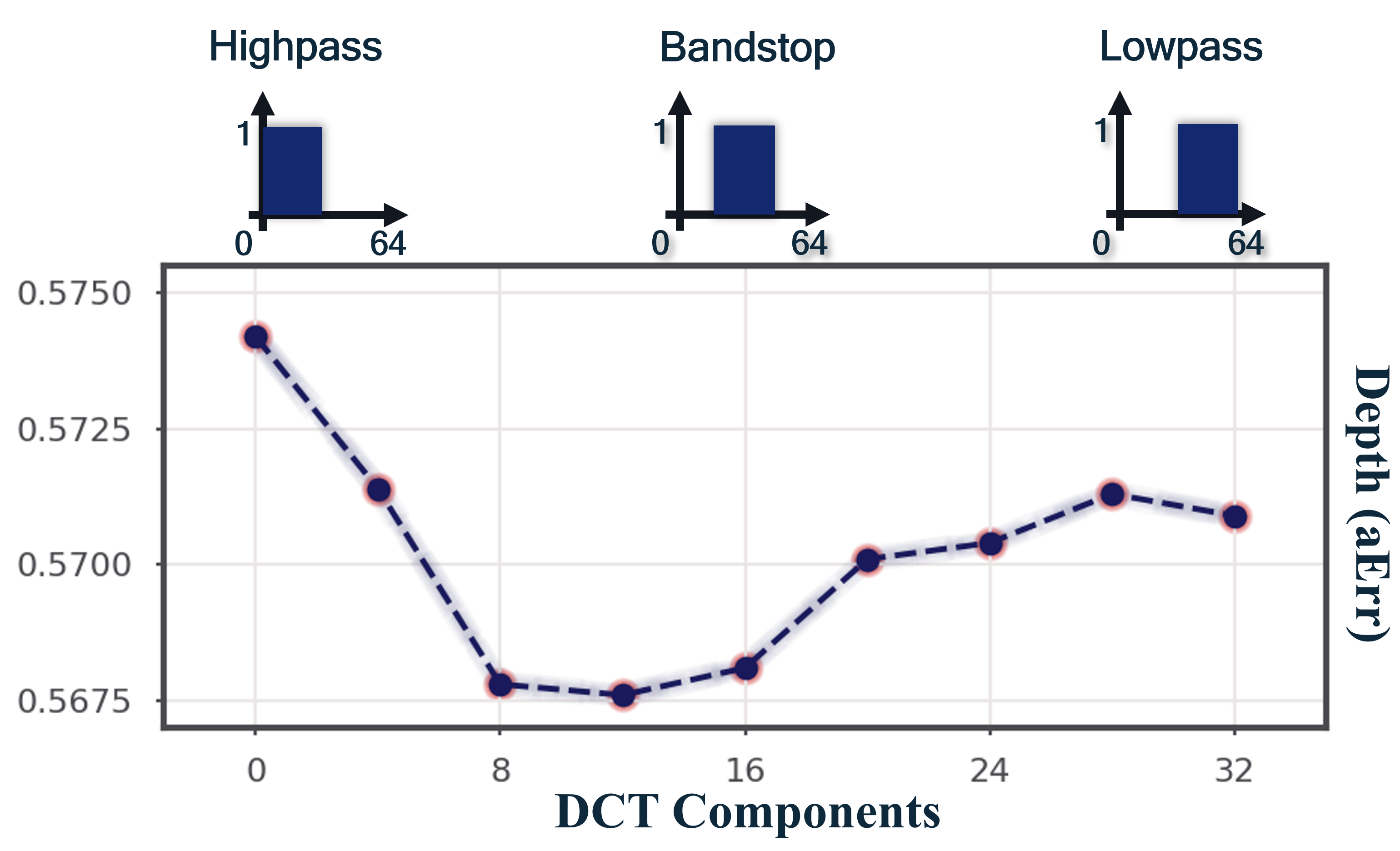}
\end{center}
\vspace{-17pt}
\caption{
Redacting various bands of frequencies to observe the performance with DejaVu loss on NYUD-v2 for depth estimation.
}
\vspace{-5pt}
\label{fig:frequency_redaction}
\end{figure}

In Fig.~\ref{fig:frequency_redaction}, we study the effect of redacting various bands of spectra in the spectral redaction (DCT) block for depth estimation. We find that the error is lowest when we use a middle band frequency for redaction. This is because most of shape information is contained in the middleband of spectra. Highband components contain grainy artifacts, and lowband components contain textures, both which are not expected to be present in the given condition (depth map).

\subsection{Extending DejaVu}

\textbf{DejaVu Text Supervision:} In Table~\ref{tab:DVTS}, we show the effect of applying DejaVu-TS, described in Section~\ref{subsec:extension}, as an additional loss. 
We observe that adding text supervision on top of the DejaVu loss can improve semantic segmentation performance.
This is because text supervision involves matching CLIP features between the original input image and the regenerated image. The CLIP model, trained on COCO, embeds additional knowledge based on textual context. Through the DejaVu regenerated image, this model is able to provide textual ques to the base network.

\begin{figure}[t]
\begin{center}
\includegraphics[width=0.95\linewidth]{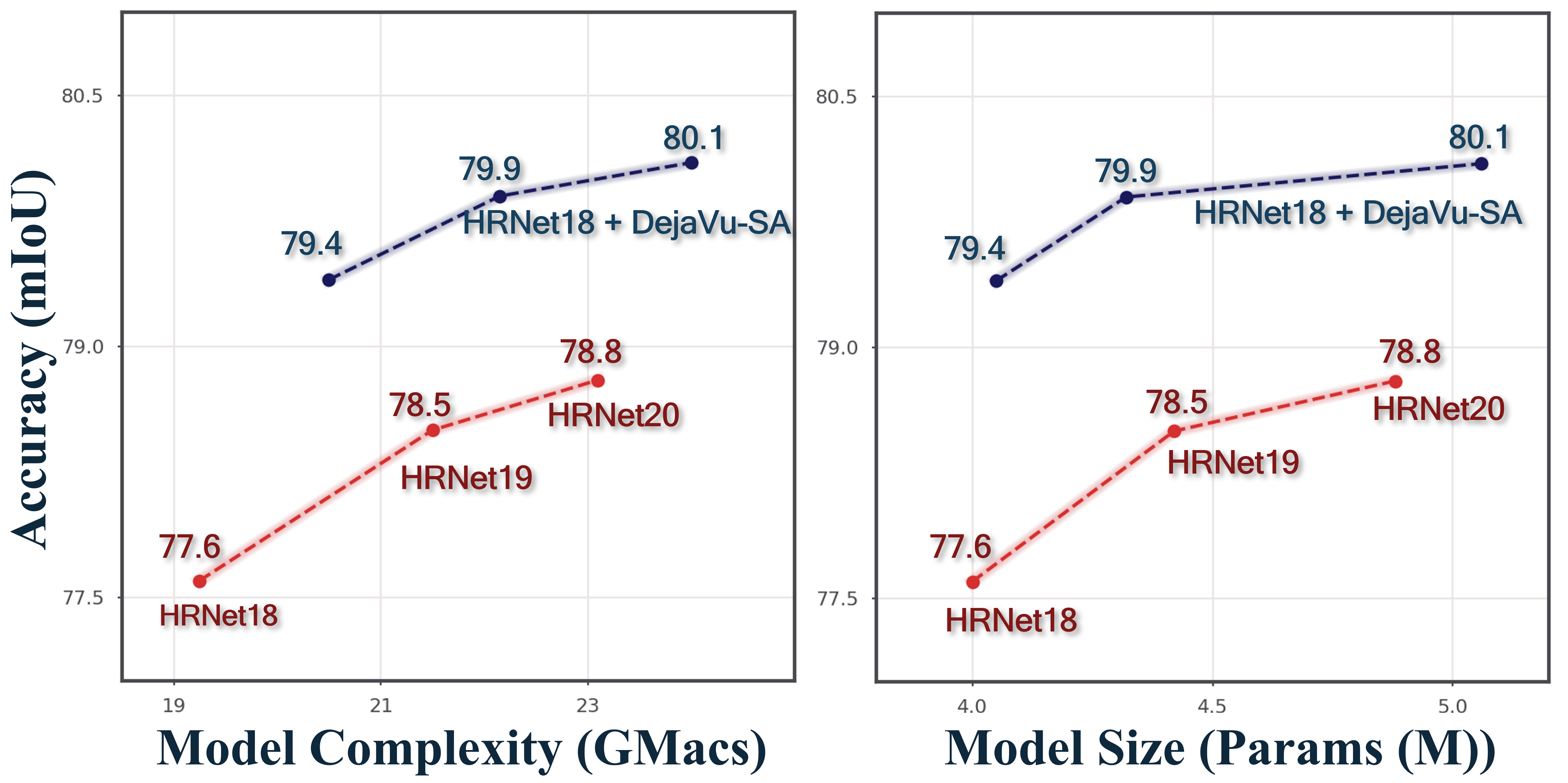}
\end{center}
\vspace{-15pt}
\caption{Performance of the proposed DejaVu-SA shared attention module, folding our regeneration operation into the network. 
DejaVu-SA module shows better performance at the same GMacs, as compared to simply scaling up the baseline architecture.
}
\vspace{-15pt}
\label{fig:dvsa_complexity}
\end{figure}

\textbf{DejaVu Shared Attention Module:} Figure~\ref{fig:dvsa_complexity} shows accuracy vs. complexity (left) and size (right) analysis for our proposed shared attention module DejaVu-SA, when evaluated on the Cityscapes semantic segmentation task. The red plot simply scales HRNet base model using different width multipliers. The blue plot scales the DejaVu-SA module by increasing its intermediate dimensions. It is important to note that HRNet18 model with DejaVu-SA can produce a higher mIoU score over HRNet20, with lower GMacs and parameters. This implies that the enhanced performance is not due to added model complexity but due to the additional embedded regeneration-based context.

\begin{table}[h]
\captionsetup{font=small, belowskip=-10pt}
\small
\centering
\setlength{\tabcolsep}{2pt}
    \centering
        \resizebox{0.40\textwidth}{!}{  
    \begin{tabular}{l|c|ccc}
    \toprule
    \textit{Method} & \textit{Redaction} & 
    \cellcolor{BurntOrange}\textit{Seg.(mIoU)}$\uparrow$ & \cellcolor{YellowGreen}\textit{Depth(aErr)}$\downarrow$ &
    \cellcolor{Cerulean}\textit{Norm(mErr)}$\downarrow$ \\
    \midrule
    Baseline & \xm & 37.25 & 59.70 & 26.30 \\
    \rowcolor{gray85}\textbf{Ours} & Spatial & \textbf{38.38} & 58.34 & 26.07 \\
    \rowcolor{gray85} \textbf{Ours} & Spectral & 38.21 & \textbf{56.76} & \textbf{25.75} \\
    \bottomrule
    \end{tabular}}
    \vspace{-5pt}
    \caption{Studying the performance of spatial v/s spectral redaction on NYUD-v2 on three tasks, using the single-task-learning setting.}
    \label{tab:space_vs_freq_nyu}
\end{table}

\section{Conclusion}
\label{sec:conclusion}
We proposed the DejaVu framework for enhancing the performance on various dense prediction tasks. DejaVu consists of adding a conditional regeneration module to reconstruct original inputs from redacted inputs and dense predictions. The reconstruction loss serves as an additional objective to produce a more generalizable dense prediction network. Additionally, we extended our framework to include an  attention mechanism, text supervision, and cycle consistency losses. We performed experiments using DejaVu on various backbones on different dense prediction tasks such as segmentation, depth estimation, and surface normal estimation. Results show that our framework produces significant improvement in performance over existing baselines both quantitatively and qualitatively. We also conducted various additional analyses and ablations to study the efficacy of different design choices in our framework. 

{\small
\bibliographystyle{ieee_fullname}
\bibliography{egbib}

\begin{thebibliography}{10}\itemsep=-1pt

\bibitem{bae2021estimating}
Gwangbin Bae, Ignas Budvytis, and Roberto Cipolla.
\newblock Estimating and exploiting the aleatoric uncertainty in surface normal
  estimation.
\newblock In {\em Proceedings of the IEEE/CVF International Conference on
  Computer Vision}, pages 13137--13146, 2021.

\bibitem{bansal2022cold}
Arpit Bansal, Eitan Borgnia, Hong-Min Chu, Jie~S Li, Hamid Kazemi, Furong
  Huang, Micah Goldblum, Jonas Geiping, and Tom Goldstein.
\newblock Cold diffusion: Inverting arbitrary image transforms without noise.
\newblock {\em arXiv preprint arXiv:2208.09392}, 2022.

\bibitem{bao2021beit}
Hangbo Bao, Li Dong, and Furu Wei.
\newblock Beit: Bert pre-training of image transformers.
\newblock {\em arXiv preprint arXiv:2106.08254}, 2021.

\bibitem{borse2021hs3}
Shubhankar Borse, Hong Cai, Yizhe Zhang, and Fatih Porikli.
\newblock Hs3: Learning with proper task complexity in hierarchically
  supervised semantic segmentation.
\newblock {\em arXiv preprint arXiv:2111.02333}, 2021.

\bibitem{borse2023x}
Shubhankar Borse, Marvin Klingner, Varun~Ravi Kumar, Hong Cai, Abdulaziz
  Almuzairee, Senthil Yogamani, and Fatih Porikli.
\newblock X-align: Cross-modal cross-view alignment for bird's-eye-view
  segmentation.
\newblock In {\em Proceedings of the IEEE/CVF Winter Conference on Applications
  of Computer Vision}, pages 3287--3297, 2023.

\bibitem{borse2022panoptic}
Shubhankar Borse, Hyojin Park, Hong Cai, Debasmit Das, Risheek Garrepalli, and
  Fatih Porikli.
\newblock Panoptic, instance and semantic relations: A relational context
  encoder to enhance panoptic segmentation.
\newblock In {\em Proceedings of the IEEE/CVF Conference on Computer Vision and
  Pattern Recognition}, pages 1269--1279, 2022.

\bibitem{borse2021inverseform}
Shubhankar Borse, Ying Wang, Yizhe Zhang, and Fatih Porikli.
\newblock Inverseform: A loss function for structured boundary-aware
  segmentation.
\newblock In {\em Proceedings of the IEEE/CVF Conference on Computer Vision and
  Pattern Recognition}, pages 5901--5911, 2021.

\bibitem{Cai2021xdistill}
Hong Cai, Janarbek Matai, Shubhankar Borse, Yizhe Zhang, Amin Ansari, and Fatih
  Porikli.
\newblock X-distill: Improving self-supervised monocular depth via cross-task
  distillation.
\newblock In {\em Proceedings of the British Machine Vision Conference}, 2022.

\bibitem{caruana1997multitask}
Rich Caruana.
\newblock Multitask learning.
\newblock {\em Machine learning}, 28(1):41--75, 1997.

\bibitem{deeplabV2}
Liang-Chieh Chen, George Papandreou, Iasonas Kokkinos, Kevin Murphy, and Alan~L
  Yuille.
\newblock {DeepLab}: Semantic image segmentation with deep convolutional nets,
  atrous convolution, and fully connected {CRF}s.
\newblock {\em PAMI}, 2018.

\bibitem{deeplabV3}
Liang-Chieh Chen, George Papandreou, Florian Schroff, and Hartwig Adam.
\newblock Rethinking atrous convolution for semantic image segmentation.
\newblock {\em arXiv:1706.05587}, 2017.

\bibitem{chen2020generative}
Mark Chen, Alec Radford, Rewon Child, Jeffrey Wu, Heewoo Jun, David Luan, and
  Ilya Sutskever.
\newblock Generative pretraining from pixels.
\newblock In {\em International conference on machine learning}, pages
  1691--1703. PMLR, 2020.

\bibitem{chen2018gradnorm}
Zhao Chen, Vijay Badrinarayanan, Chen-Yu Lee, and Andrew Rabinovich.
\newblock Gradnorm: Gradient normalization for adaptive loss balancing in deep
  multitask networks.
\newblock In {\em International conference on machine learning}, pages
  794--803. PMLR, 2018.

\bibitem{chen2022vitadapter}
Zhe Chen, Yuchen Duan, Wenhai Wang, Junjun He, Tong Lu, Jifeng Dai, and Yu
  Qiao.
\newblock Vision transformer adapter for dense predictions.
\newblock {\em arXiv preprint arXiv:2205.08534}, 2022.

\bibitem{cheng2020panoptic}
Bowen Cheng, Maxwell~D Collins, Yukun Zhu, Ting Liu, Thomas~S Huang, Hartwig
  Adam, and Liang-Chieh Chen.
\newblock {Panoptic-DeepLab}: A simple, strong, and fast baseline for bottom-up
  panoptic segmentation.
\newblock In {\em CVPR}, 2020.

\bibitem{cheng2022masked}
Bowen Cheng, Ishan Misra, Alexander~G Schwing, Alexander Kirillov, and Rohit
  Girdhar.
\newblock Masked-attention mask transformer for universal image segmentation.
\newblock In {\em Proceedings of the IEEE/CVF Conference on Computer Vision and
  Pattern Recognition}, pages 1290--1299, 2022.

\bibitem{cheng2021maskformer}
Bowen Cheng, Alexander~G. Schwing, and Alexander Kirillov.
\newblock Per-pixel classification is not all you need for semantic
  segmentation.
\newblock In {\em NeurIPS}, 2021.

\bibitem{cordts2016cityscapes}
Marius Cordts, Mohamed Omran, Sebastian Ramos, Timo Rehfeld, Markus Enzweiler,
  Rodrigo Benenson, Uwe Franke, Stefan Roth, and Bernt Schiele.
\newblock The {Cityscapes} dataset for semantic urban scene understanding.
\newblock In {\em CVPR}, 2016.

\bibitem{di2021pixel}
Giancarlo Di~Biase, Hermann Blum, Roland Siegwart, and Cesar Cadena.
\newblock Pixel-wise anomaly detection in complex driving scenes.
\newblock In {\em Proceedings of the IEEE/CVF conference on computer vision and
  pattern recognition}, pages 16918--16927, 2021.

\bibitem{do2020surface}
Tien Do, Khiem Vuong, Stergios~I Roumeliotis, and Hyun~Soo Park.
\newblock Surface normal estimation of tilted images via spatial rectifier.
\newblock In {\em European Conference on Computer Vision}, pages 265--280.
  Springer, 2020.

\bibitem{dosovitskiy2020image}
Alexey Dosovitskiy, Lucas Beyer, Alexander Kolesnikov, Dirk Weissenborn,
  Xiaohua Zhai, Thomas Unterthiner, Mostafa Dehghani, Matthias Minderer, Georg
  Heigold, Sylvain Gelly, et~al.
\newblock An image is worth 16x16 words: Transformers for image recognition at
  scale.
\newblock {\em arXiv preprint arXiv:2010.11929}, 2020.

\bibitem{eigen2014depth}
David Eigen, Christian Puhrsch, and Rob Fergus.
\newblock Depth map prediction from a single image using a multi-scale deep
  network.
\newblock {\em Advances in neural information processing systems}, 27, 2014.

\bibitem{fang2022eva}
Yuxin Fang, Wen Wang, Binhui Xie, Quan Sun, Ledell Wu, Xinggang Wang, Tiejun
  Huang, Xinlong Wang, and Yue Cao.
\newblock Eva: Exploring the limits of masked visual representation learning at
  scale.
\newblock {\em arXiv preprint arXiv:2211.07636}, 2022.

\bibitem{fu2019dual}
Jun Fu, Jing Liu, Haijie Tian, Yong Li, Yongjun Bao, Zhiwei Fang, and Hanqing
  Lu.
\newblock Dual attention network for scene segmentation.
\newblock In {\em CVPR}, 2019.

\bibitem{garrepalli2022oracle}
Risheek Garrepalli.
\newblock Oracle analysis of representations for deep open set detection.
\newblock {\em arXiv preprint arXiv:2209.11350}, 2022.

\bibitem{geiger2013vision}
Andreas Geiger, Philip Lenz, Christoph Stiller, and Raquel Urtasun.
\newblock Vision meets robotics: The kitti dataset.
\newblock {\em The International Journal of Robotics Research},
  32(11):1231--1237, 2013.

\bibitem{godard2017unsupervised}
Cl{\'e}ment Godard, Oisin Mac~Aodha, and Gabriel~J Brostow.
\newblock Unsupervised monocular depth estimation with left-right consistency.
\newblock In {\em Proceedings of the IEEE conference on computer vision and
  pattern recognition}, pages 270--279, 2017.

\bibitem{godard2019digging}
Cl{\'e}ment Godard, Oisin Mac~Aodha, Michael Firman, and Gabriel~J Brostow.
\newblock Digging into self-supervised monocular depth estimation.
\newblock In {\em Proceedings of the IEEE/CVF International Conference on
  Computer Vision}, pages 3828--3838, 2019.

\bibitem{guizilini20203d}
Vitor Guizilini, Rares Ambrus, Sudeep Pillai, Allan Raventos, and Adrien
  Gaidon.
\newblock 3d packing for self-supervised monocular depth estimation.
\newblock In {\em Proceedings of the IEEE/CVF Conference on Computer Vision and
  Pattern Recognition}, pages 2485--2494, 2020.

\bibitem{guizilini2019semantically}
Vitor Guizilini, Rui Hou, Jie Li, Rares Ambrus, and Adrien Gaidon.
\newblock Semantically-guided representation learning for self-supervised
  monocular depth.
\newblock In {\em International Conference on Learning Representations}, 2019.

\bibitem{he2022masked}
Kaiming He, Xinlei Chen, Saining Xie, Yanghao Li, Piotr Doll{\'a}r, and Ross
  Girshick.
\newblock Masked autoencoders are scalable vision learners.
\newblock In {\em Proceedings of the IEEE/CVF Conference on Computer Vision and
  Pattern Recognition}, pages 16000--16009, 2022.

\bibitem{hinton1993autoencoders}
Geoffrey~E Hinton and Richard Zemel.
\newblock Autoencoders, minimum description length and helmholtz free energy.
\newblock {\em Advances in neural information processing systems}, 6, 1993.

\bibitem{huang2019ccnet}
Zilong Huang, Xinggang Wang, Lichao Huang, Chang Huang, Yunchao Wei, and Wenyu
  Liu.
\newblock {CCNet}: Criss-cross attention for semantic segmentation.
\newblock In {\em ICCV}, 2019.

\bibitem{jain2021semask}
Jitesh Jain, Anukriti Singh, Nikita Orlov, Zilong Huang, Jiachen Li, Steven
  Walton, and Humphrey Shi.
\newblock Semask: Semantically masking transformer backbones for effective
  semantic segmentation.
\newblock {\em arXiv}, 2021.

\bibitem{jiang2020dipe}
Hualie Jiang, Laiyan Ding, Zhenglong Sun, and Rui Huang.
\newblock Dipe: Deeper into photometric errors for unsupervised learning of
  depth and ego-motion from monocular videos.
\newblock In {\em 2020 IEEE/RSJ International Conference on Intelligent Robots
  and Systems (IROS)}, pages 10061--10067. IEEE, 2020.

\bibitem{karnam2020self}
Srivallabha Karnam.
\newblock {\em Self-Supervised Learning for Segmentation using Image
  Reconstruction}.
\newblock Rochester Institute of Technology, 2020.

\bibitem{karras2020analyzing}
Tero Karras, Samuli Laine, Miika Aittala, Janne Hellsten, Jaakko Lehtinen, and
  Timo Aila.
\newblock Analyzing and improving the image quality of stylegan.
\newblock In {\em Proceedings of the IEEE/CVF conference on computer vision and
  pattern recognition}, pages 8110--8119, 2020.

\bibitem{kendall2017multi}
Alex Kendall, Yarin Gal, and Roberto Cipolla.
\newblock Multi-task learning using uncertainty to weigh losses for scene
  geometry and semantics.
\newblock In {\em CVPR}, 2018.

\bibitem{kirillov2019panopticfpn}
Alexander Kirillov, Ross Girshick, Kaiming He, and Piotr Doll{\'a}r.
\newblock Panoptic feature pyramid networks.
\newblock In {\em CVPR}, 2019.

\bibitem{kirillov2019panoptic}
Alexander Kirillov, Ross Girshick, Kaiming He, and Piotr Doll{\'a}r.
\newblock Panoptic feature pyramid networks.
\newblock In {\em Proceedings of the IEEE/CVF conference on computer vision and
  pattern recognition}, pages 6399--6408, 2019.

\bibitem{klingner2023x}
Marvin Klingner, Shubhankar Borse, Varun~Ravi Kumar, Behnaz Rezaei, Venkatraman
  Narayanan, Senthil Yogamani, and Fatih Porikli.
\newblock X$^3$kd: Knowledge distillation across modalities, tasks and stages
  for multi-camera 3d object detection.
\newblock {\em arXiv preprint arXiv:2303.02203}, 2023.

\bibitem{klingner2020self}
Marvin Klingner, Jan-Aike Term{\"o}hlen, Jonas Mikolajczyk, and Tim
  Fingscheidt.
\newblock Self-supervised monocular depth estimation: Solving the dynamic
  object problem by semantic guidance.
\newblock In {\em European Conference on Computer Vision}, pages 582--600.
  Springer, 2020.

\bibitem{li2022learning}
Wei-Hong Li, Xialei Liu, and Hakan Bilen.
\newblock Learning multiple dense prediction tasks from partially annotated
  data.
\newblock In {\em Proceedings of the IEEE/CVF Conference on Computer Vision and
  Pattern Recognition}, pages 18879--18889, 2022.

\bibitem{li2021fully}
Yanwei Li, Hengshuang Zhao, Xiaojuan Qi, Yukang Chen, Lu Qi, Liwei Wang, Zeming
  Li, Jian Sun, and Jiaya Jia.
\newblock Fully convolutional networks for panoptic segmentation with
  point-based supervision.
\newblock {\em arXiv preprint arXiv:2108.07682}, 2021.

\bibitem{liao2019spherical}
Shuai Liao, Efstratios Gavves, and Cees~GM Snoek.
\newblock Spherical regression: Learning viewpoints, surface normals and 3d
  rotations on n-spheres.
\newblock In {\em Proceedings of the IEEE/CVF Conference on Computer Vision and
  Pattern Recognition}, pages 9759--9767, 2019.

\bibitem{lin2017focal}
Tsung-Yi Lin, Priya Goyal, Ross Girshick, Kaiming He, and Piotr Doll{\'a}r.
\newblock Focal loss for dense object detection.
\newblock In {\em ICCV}, 2017.

\bibitem{lin2014microsoft}
Tsung-Yi Lin, Michael Maire, Serge Belongie, James Hays, Pietro Perona, Deva
  Ramanan, Piotr Doll{\'a}r, and C~Lawrence Zitnick.
\newblock Microsoft coco: Common objects in context.
\newblock In {\em European conference on computer vision}, pages 740--755.
  Springer, 2014.

\bibitem{pmlr-v80-liu18e}
Si Liu, Risheek Garrepalli, Thomas Dietterich, Alan Fern, and Dan Hendrycks.
\newblock Open category detection with {PAC} guarantees.
\newblock In Jennifer Dy and Andreas Krause, editors, {\em Proceedings of the
  35th International Conference on Machine Learning}, volume~80 of {\em
  Proceedings of Machine Learning Research}, pages 3169--3178. PMLR, 10--15 Jul
  2018.

\bibitem{liu2022pac}
Si Liu, Risheek Garrepalli, Dan Hendrycks, Alan Fern, Debashis Mondal, and
  Thomas~G Dietterich.
\newblock Pac guarantees and effective algorithms for detecting novel
  categories.
\newblock {\em J. Mach. Learn. Res.}, 23:44--1, 2022.

\bibitem{liu2019end}
Shikun Liu, Edward Johns, and Andrew~J Davison.
\newblock End-to-end multi-task learning with attention.
\newblock In {\em Proceedings of the IEEE/CVF conference on computer vision and
  pattern recognition}, pages 1871--1880, 2019.

\bibitem{liu2019learning}
Xihui Liu, Guojun Yin, Jing Shao, Xiaogang Wang, et~al.
\newblock Learning to predict layout-to-image conditional convolutions for
  semantic image synthesis.
\newblock {\em Advances in Neural Information Processing Systems}, 32, 2019.

\bibitem{liu2021swin}
Ze Liu, Yutong Lin, Yue Cao, Han Hu, Yixuan Wei, Zheng Zhang, Stephen Lin, and
  Baining Guo.
\newblock Swin transformer: Hierarchical vision transformer using shifted
  windows.
\newblock {\em arXiv:2103.14030}, 2021.

\bibitem{long2015fully}
Jonathan Long, Evan Shelhamer, and Trevor Darrell.
\newblock Fully convolutional networks for semantic segmentation.
\newblock In {\em CVPR}, 2015.

\bibitem{lu2020depth}
Kaiyue Lu, Nick Barnes, Saeed Anwar, and Liang Zheng.
\newblock From depth what can you see? depth completion via auxiliary image
  reconstruction.
\newblock In {\em Proceedings of the IEEE/CVF Conference on Computer Vision and
  Pattern Recognition}, pages 11306--11315, 2020.

\bibitem{nalisnick2019hybrid}
Eric Nalisnick, Akihiro Matsukawa, Yee~Whye Teh, Dilan Gorur, and Balaji
  Lakshminarayanan.
\newblock Hybrid models with deep and invertible features.
\newblock In {\em International Conference on Machine Learning}, pages
  4723--4732. PMLR, 2019.

\bibitem{silberman2012nyu}
Pushmeet~Kohli Nathan~Silberman, Derek~Hoiem and Rob Fergus.
\newblock Indoor segmentation and support inference from rgbd images.
\newblock In {\em ECCV}, 2012.

\bibitem{park2019semantic}
Taesung Park, Ming-Yu Liu, Ting-Chun Wang, and Jun-Yan Zhu.
\newblock Semantic image synthesis with spatially-adaptive normalization.
\newblock In {\em Proceedings of the IEEE/CVF conference on computer vision and
  pattern recognition}, pages 2337--2346, 2019.

\bibitem{pathak2016context}
Deepak Pathak, Philipp Krahenbuhl, Jeff Donahue, Trevor Darrell, and Alexei~A
  Efros.
\newblock Context encoders: Feature learning by inpainting.
\newblock In {\em Proceedings of the IEEE conference on computer vision and
  pattern recognition}, pages 2536--2544, 2016.

\bibitem{patil2020don}
Vaishakh Patil, Wouter Van~Gansbeke, Dengxin Dai, and Luc Van~Gool.
\newblock Don’t forget the past: Recurrent depth estimation from monocular
  video.
\newblock {\em IEEE Robotics and Automation Letters}, 5(4):6813--6820, 2020.

\bibitem{qi2018geonet}
Xiaojuan Qi, Renjie Liao, Zhengzhe Liu, Raquel Urtasun, and Jiaya Jia.
\newblock Geonet: Geometric neural network for joint depth and surface normal
  estimation.
\newblock In {\em Proceedings of the IEEE Conference on Computer Vision and
  Pattern Recognition}, pages 283--291, 2018.

\bibitem{qi2020geonet++}
Xiaojuan Qi, Zhengzhe Liu, Renjie Liao, Philip~HS Torr, Raquel Urtasun, and
  Jiaya Jia.
\newblock Geonet++: Iterative geometric neural network with edge-aware
  refinement for joint depth and surface normal estimation.
\newblock {\em IEEE Transactions on Pattern Analysis and Machine Intelligence},
  2020.

\bibitem{radford2021learning}
Alec Radford, Jong~Wook Kim, Chris Hallacy, Aditya Ramesh, Gabriel Goh,
  Sandhini Agarwal, Girish Sastry, Amanda Askell, Pamela Mishkin, Jack Clark,
  et~al.
\newblock Learning transferable visual models from natural language
  supervision.
\newblock In {\em International Conference on Machine Learning}, pages
  8748--8763. PMLR, 2021.

\bibitem{rao2022denseclip}
Yongming Rao, Wenliang Zhao, Guangyi Chen, Yansong Tang, Zheng Zhu, Guan Huang,
  Jie Zhou, and Jiwen Lu.
\newblock Denseclip: Language-guided dense prediction with context-aware
  prompting.
\newblock In {\em Proceedings of the IEEE/CVF Conference on Computer Vision and
  Pattern Recognition}, pages 18082--18091, 2022.

\bibitem{rombach2022high}
Robin Rombach, Andreas Blattmann, Dominik Lorenz, Patrick Esser, and Bj{\"o}rn
  Ommer.
\newblock High-resolution image synthesis with latent diffusion models.
\newblock In {\em Proceedings of the IEEE/CVF Conference on Computer Vision and
  Pattern Recognition}, pages 10684--10695, 2022.

\bibitem{sanakoyeu2018style}
Artsiom Sanakoyeu, Dmytro Kotovenko, Sabine Lang, and Bjorn Ommer.
\newblock A style-aware content loss for real-time hd style transfer.
\newblock In {\em proceedings of the European conference on computer vision
  (ECCV)}, pages 698--714, 2018.

\bibitem{sener2018multi}
Ozan Sener and Vladlen Koltun.
\newblock Multi-task learning as multi-objective optimization.
\newblock {\em Advances in neural information processing systems}, 31, 2018.

\bibitem{strudel2021segmenter}
Robin Strudel, Ricardo Garcia, Ivan Laptev, and Cordelia Schmid.
\newblock Segmenter: Transformer for semantic segmentation.
\newblock In {\em ICCV}, 2021.

\bibitem{su2022towards}
Weijie Su, Xizhou Zhu, Chenxin Tao, Lewei Lu, Bin Li, Gao Huang, Yu Qiao,
  Xiaogang Wang, Jie Zhou, and Jifeng Dai.
\newblock Towards all-in-one pre-training via maximizing multi-modal mutual
  information.
\newblock {\em arXiv preprint arXiv:2211.09807}, 2022.

\bibitem{sun2020conditional}
Xin Sun, Zhenning Yang, Chi Zhang, Keck-Voon Ling, and Guohao Peng.
\newblock Conditional gaussian distribution learning for open set recognition.
\newblock In {\em Proceedings of the IEEE/CVF Conference on Computer Vision and
  Pattern Recognition}, pages 13480--13489, 2020.

\bibitem{tao2020hierarchical}
Andrew Tao, Karan Sapra, and Bryan Catanzaro.
\newblock Hierarchical multi-scale attention for semantic segmentation.
\newblock {\em arXiv:2005.10821}, 2020.

\bibitem{ulyanov2017improved}
Dmitry Ulyanov, Andrea Vedaldi, and Victor Lempitsky.
\newblock Improved texture networks: Maximizing quality and diversity in
  feed-forward stylization and texture synthesis.
\newblock In {\em Proceedings of the IEEE conference on computer vision and
  pattern recognition}, pages 6924--6932, 2017.

\bibitem{vincent2008extracting}
Pascal Vincent, Hugo Larochelle, Yoshua Bengio, and Pierre-Antoine Manzagol.
\newblock Extracting and composing robust features with denoising autoencoders.
\newblock In {\em Proceedings of the 25th international conference on Machine
  learning}, pages 1096--1103, 2008.

\bibitem{vincent2010stacked}
Pascal Vincent, Hugo Larochelle, Isabelle Lajoie, Yoshua Bengio, Pierre-Antoine
  Manzagol, and L{\'e}on Bottou.
\newblock Stacked denoising autoencoders: Learning useful representations in a
  deep network with a local denoising criterion.
\newblock {\em Journal of machine learning research}, 11(12), 2010.

\bibitem{wang2021max}
Huiyu Wang, Yukun Zhu, Hartwig Adam, Alan Yuille, and Liang-Chieh Chen.
\newblock {MaX-DeepLab}: End-to-end panoptic segmentation with mask
  transformers.
\newblock In {\em CVPR}, 2021.

\bibitem{WangSCJDZLMTWLX19hrnet}
Jingdong Wang, Ke Sun, Tianheng Cheng, Borui Jiang, Chaorui Deng, Yang Zhao,
  Dong Liu, Yadong Mu, Mingkui Tan, Xinggang Wang, Wenyu Liu, and Bin Xiao.
\newblock Deep high-resolution representation learning for visual recognition.
\newblock {\em PAMI}, 2019.

\bibitem{wang2018high}
Ting-Chun Wang, Ming-Yu Liu, Jun-Yan Zhu, Andrew Tao, Jan Kautz, and Bryan
  Catanzaro.
\newblock High-resolution image synthesis and semantic manipulation with
  conditional gans.
\newblock In {\em Proceedings of the IEEE conference on computer vision and
  pattern recognition}, pages 8798--8807, 2018.

\bibitem{wang2022image}
Wenhui Wang, Hangbo Bao, Li Dong, Johan Bjorck, Zhiliang Peng, Qiang Liu, Kriti
  Aggarwal, Owais~Khan Mohammed, Saksham Singhal, Subhojit Som, et~al.
\newblock Image as a foreign language: Beit pretraining for all vision and
  vision-language tasks.
\newblock {\em arXiv preprint arXiv:2208.10442}, 2022.

\bibitem{wang2022internimage}
Wenhai Wang, Jifeng Dai, Zhe Chen, Zhenhang Huang, Zhiqi Li, Xizhou Zhu,
  Xiaowei Hu, Tong Lu, Lewei Lu, Hongsheng Li, et~al.
\newblock Internimage: Exploring large-scale vision foundation models with
  deformable convolutions.
\newblock {\em arXiv preprint arXiv:2211.05778}, 2022.

\bibitem{wang2018non}
Xiaolong Wang, Ross Girshick, Abhinav Gupta, and Kaiming He.
\newblock Non-local neural networks.
\newblock In {\em CVPR}, 2018.

\bibitem{wang2019object}
Zian Wang, David Acuna, Huan Ling, Amlan Kar, and Sanja Fidler.
\newblock Object instance annotation with deep extreme level set evolution.
\newblock In {\em CVPR}, 2019.

\bibitem{xia2017w}
Xide Xia and Brian Kulis.
\newblock W-net: A deep model for fully unsupervised image segmentation.
\newblock {\em arXiv preprint arXiv:1711.08506}, 2017.

\bibitem{xiao2018unified}
Tete Xiao, Yingcheng Liu, Bolei Zhou, Yuning Jiang, and Jian Sun.
\newblock Unified perceptual parsing for scene understanding.
\newblock In {\em ECCV}, 2018.

\bibitem{xie2021segformer}
Enze Xie, Wenhai Wang, Zhiding Yu, Anima Anandkumar, Jose~M Alvarez, and Ping
  Luo.
\newblock Segformer: Simple and efficient design for semantic segmentation with
  transformers.
\newblock In {\em NeurIPS}, 2021.

\bibitem{xiong19upsnet}
Yuwen Xiong, Renjie Liao, Hengshuang Zhao, Rui Hu, Min Bai, Ersin Yumer, and
  Raquel Urtasun.
\newblock Upsnet: A unified panoptic segmentation network.
\newblock In {\em CVPR}, 2019.

\bibitem{yang2020label}
Jinyu Yang, Weizhi An, Sheng Wang, Xinliang Zhu, Chaochao Yan, and Junzhou
  Huang.
\newblock Label-driven reconstruction for domain adaptation in semantic
  segmentation.
\newblock In {\em European conference on computer vision}, pages 480--498.
  Springer, 2020.

\bibitem{yu2022metaformer}
Weihao Yu, Mi Luo, Pan Zhou, Chenyang Si, Yichen Zhou, Xinchao Wang, Jiashi
  Feng, and Shuicheng Yan.
\newblock Metaformer is actually what you need for vision.
\newblock In {\em Proceedings of the IEEE/CVF Conference on Computer Vision and
  Pattern Recognition}, pages 10819--10829, 2022.

\bibitem{yuan2020object}
Yuhui Yuan, Xilin Chen, and Jingdong Wang.
\newblock Object-contextual representations for semantic segmentation.
\newblock In {\em European conference on computer vision}, pages 173--190.
  Springer, 2020.

\bibitem{yuan2018ocnet}
Yuhui Yuan, Lang Huang, Jianyuan Guo, Chao Zhang, Xilin Chen, and Jingdong
  Wang.
\newblock {OCNet}: Object context for semantic segmentation.
\newblock {\em IJCV}, 2021.

\bibitem{zhang2016colorful}
Richard Zhang, Phillip Isola, and Alexei~A Efros.
\newblock Colorful image colorization.
\newblock In {\em European conference on computer vision}, pages 649--666.
  Springer, 2016.

\bibitem{zhang2018unreasonable}
Richard Zhang, Phillip Isola, Alexei~A Efros, Eli Shechtman, and Oliver Wang.
\newblock The unreasonable effectiveness of deep features as a perceptual
  metric.
\newblock In {\em Proceedings of the IEEE conference on computer vision and
  pattern recognition}, pages 586--595, 2018.

\bibitem{zhang2021knet}
Wenwei Zhang, Jiangmiao Pang, Kai Chen, and Chen~Change Loy.
\newblock K-net: Towards unified image segmentation.
\newblock In {\em NeurIPS}, 2021.

\bibitem{zhang2022auxadapt}
Yizhe Zhang, Shubhankar Borse, Hong Cai, and Fatih Porikli.
\newblock Auxadapt: Stable and efficient test-time adaptation for temporally
  consistent video semantic segmentation.
\newblock In {\em Proceedings of the IEEE/CVF Winter Conference on Applications
  of Computer Vision}, pages 2339--2348, 2022.

\bibitem{zhang2022perceptual}
Yizhe Zhang, Shubhankar Borse, Hong Cai, Ying Wang, Ning Bi, Xiaoyun Jiang, and
  Fatih Porikli.
\newblock Perceptual consistency in video segmentation.
\newblock In {\em Proceedings of the IEEE/CVF Winter Conference on Applications
  of Computer Vision}, pages 2564--2573, 2022.

\bibitem{zhang2016augmenting}
Yuting Zhang, Kibok Lee, and Honglak Lee.
\newblock Augmenting supervised neural networks with unsupervised objectives
  for large-scale image classification.
\newblock In {\em International conference on machine learning}, pages
  612--621. PMLR, 2016.

\bibitem{zhao2017pspnet}
Hengshuang Zhao, Jianping Shi, Xiaojuan Qi, Xiaogang Wang, and Jiaya Jia.
\newblock Pyramid scene parsing network.
\newblock In {\em CVPR}, 2017.

\bibitem{zheng2021rethinking}
Sixiao Zheng, Jiachen Lu, Hengshuang Zhao, Xiatian Zhu, Zekun Luo, Yabiao Wang,
  Yanwei Fu, Jianfeng Feng, Tao Xiang, Philip~HS Torr, et~al.
\newblock Rethinking semantic segmentation from a sequence-to-sequence
  perspective with transformers.
\newblock In {\em Proceedings of the IEEE/CVF conference on computer vision and
  pattern recognition}, pages 6881--6890, 2021.

\bibitem{zhou2017scene}
Bolei Zhou, Hang Zhao, Xavier Puig, Sanja Fidler, Adela Barriuso, and Antonio
  Torralba.
\newblock Scene parsing through ade20k dataset.
\newblock In {\em Proceedings of the IEEE conference on computer vision and
  pattern recognition}, pages 633--641, 2017.

\bibitem{zhou2017unsupervised}
Tinghui Zhou, Matthew Brown, Noah Snavely, and David~G Lowe.
\newblock Unsupervised learning of depth and ego-motion from video.
\newblock In {\em Proceedings of the IEEE conference on computer vision and
  pattern recognition}, pages 1851--1858, 2017.

\end{thebibliography}
}

\end{document}